\documentclass[10pt]{article}
\usepackage[margin=1in]{geometry}
\usepackage{times}
\usepackage[numbers,sort&compress]{natbib}
\usepackage{hyperref}
\usepackage{url}
\usepackage{graphicx}
\usepackage{subcaption}
\usepackage{multirow}
\usepackage{booktabs}
\usepackage{amsmath}
\usepackage{amssymb}
\usepackage{xspace}
\usepackage{xcolor}
\usepackage{pifont}
\newcommand{\cmark}{\ding{51}}
\newcommand{\xmark}{\ding{55}}
\newcommand{\bench}{\textsc{NegT2IBench}\xspace}

\title{
\bench: When Negation Changes the Picture. A Polarity Benchmark for Text-to-Image Models
}
\author{
Omar Elfatairy\textsuperscript{*} \quad 
María A. Bravo\textsuperscript{*} \quad 
Jessica Bader\textsuperscript{\textdagger} \quad 
Zeynep Akata\\[0.75em]
\normalsize Technical University of Munich,  \quad 
\normalsize Helmholtz Munich,  \quad 
\normalsize Munich Center for Machine Learning \\[0.5em]
\small Corresponding authors: \texttt{omar.elfatairy@tum.de},  \quad \texttt{maria.bravo@tum.de} \\ [0.5em]
\small Code: \hyperlink{https://github.com/Fatairy/NegT2IBench}{\texttt{https://github.com/Fatairy/NegT2IBench}}}
\date{}
\begin{document}

\maketitle

\begingroup
\renewcommand{\thefootnote}{*}
\footnotetext{Equal contribution.}
\endgroup

\begingroup
\renewcommand{\thefootnote}{\textdagger}
\footnotetext{Work done while at Technical University of Munich and Helmholtz Munich.}
\endgroup

\begin{abstract}
Text-to-image (T2I) models are judged by benchmarks that measure whether requested content appears, but these benchmarks largely overlook the complementary ability to satisfy negated constraints, for example, generating ``a non-red cup.'' Measuring negation raises challenges not faced by affirmation-based benchmarks and requires careful prompt and evaluation design.
We introduce \bench, a benchmark of 4,800 prompts covering two attribute types and four relation categories. Prompts are organized by \textit{polarity}: the number of \textit{positive} statements that must hold and \textit{negated} statements that must not, each ranging from 0 to 2. Varying the two independently separates the effect of negation from the effect of prompt complexity.
Our detector-based scoring is reproducible, auditable, and pinpoints which requirement failed. On 600 images with three-annotator labels, it agrees with humans as closely as vision-language judges up to $30\times$ larger, while using only a fraction of their GPU memory.
Across eleven T2I models and 211,200 images, nine score lower on a single negated statement than on a single positive one. Per-statement scoring reveals that the loss is largest for color and near zero for proximity, and that 41.5\% of failed statements render exactly what the prompt forbids.
Rendering what a prompt asks for and withholding what it forbids are distinct capabilities that an aggregate compositional score cannot distinguish. \bench measures the latter directly, providing a controlled testbed for diagnosing negation failures and developing methods to overcome them.
\end{abstract}

\definecolor{check_green}{HTML}{1c8e3e}
\definecolor{x_red}{HTML}{d93025}

\begin{figure}[t]
\centering
\includegraphics[width=\linewidth]{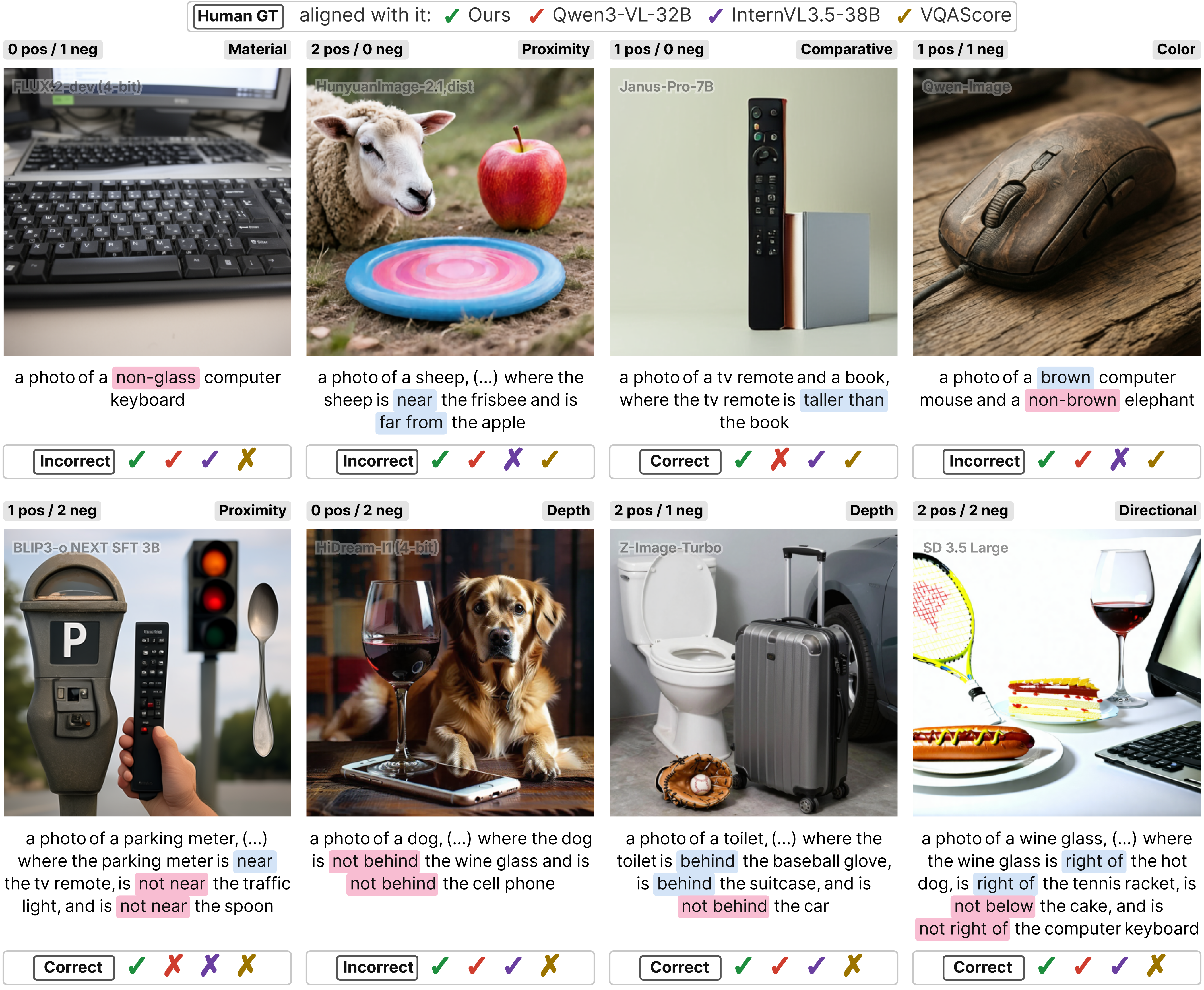}

\caption{\bench prompts mix positive (blue) and negated (pink) statements, spanning six categories (above image, right) and organized by the number of each statement (above image, left). Images come from the benchmarked T2I models. The box under each image gives the human verdict; marks show whether each evaluator agrees (\cmark) or disagrees (\xmark) with it.}
\label{fig:teaser}
\end{figure}

\section{Introduction}
\label{sec:intro}
While text-to-image (T2I) models now render increasingly complex prompts accurately enough for professional~\citep{Shabani2022HouseDiffusionVF} and creative~\citep{Ko2022LargescaleTG} applications, evaluating whether they respect what a prompt \textit{forbids} is surprisingly difficult and underexplored. The frontier is judged by benchmarks like GenEval~\citep{geneval_2023ghosh}, T2I-CompBench~\citep{t2icompbench_2023huang}, and DPG-Bench~\citep{dpgbench_2024hu}, which focus on challenging, dense prompts with multiple components that must each be correct. Though these benchmarks cover many important prompt-following aspects, they largely overlook an important capability: \textit{negation}.

Measuring negation raises challenges not faced by affirmation-based benchmarks, making it more nuanced than simply adding negated prompts to existing benchmarks. First, success can be accidental: cups rarely fly, so ``a cup that is not flying'' can produce a valid image even if \textit{not} was ignored. Second, if object presence is not verified, a negated attribute can be satisfied by omitting the object rather than  the attribute: ``a non-wooden bench'' may simply omit the bench. Third, negation inherits the difficulty of dense prompts; adding objects, like ``a red cup'' vs. ``a red cup with no spoon'' introduces multiple potential failures, so failures from negation must be disentangled from those from prompt complexity. Fourth, the evaluator can also introduce errors: when checking for forbidden elements, vision-language models may report absent content~\citep{pope_2023li} or hallucinate~\citep{finer_2026xiao}, making evaluator errors difficult to distinguish from image-generation errors.

We introduce \bench, a benchmark for measuring negation in text-to-image generation, designed around these four challenges. To reduce accidental success, an LLM~\citep{qwen3_2025qwenteam} selects plausible attributes and relations, and fixed templates keep wording consistent across conditions. To prevent success by omission, evaluation requires every named object to appear: i.e., ``a non-wooden bench'' still requires a bench. To isolate negation from prompt complexity, prompts are organized by \textit{polarity}, the number of \textit{positive} statements that must hold and \textit{negated} statements that must not; prompts with the same statement count contain the same object count and differ only in polarity. Finally, to limit evaluator error, each requirement is checked by a dedicated perception component rather than by an LLM judge, revealing which requirement failed. \bench covers two attribute categories (color and material) and four relation categories (directional, proximity, comparative, and depth).

We evaluate eleven T2I models on the full corpus, four samples per prompt totaling 211,200 images (Section~\ref{sec:experiments}). Nine of eleven models score lower on a single negated statement than on a single positive one, by up to $43.8\%$ points. The drop differs by category: color falls from $85.0\%$ to $31.3\%$, while proximity barely moves, $45.8\%$ vs. $43.2\%$. At the highest tested complexity, no model exceeds $8.4\%$. More strikingly, $41.5\%$ of failed statements are negated violations: the image shows the attribute or relation the prompt forbids. Ranking also changes with polarity: FLUX.2-dev leads on the single positive statement and Janus-Pro-7B on the negated one, so a single score hides the effect. Replacing SD~1.5's text encoder with negation-aware NegationCLIP~\citep{negationclip_2025park} improves accuracy only from $14.5\%$ to $14.9\%$. Negation is therefore a failure mode of its own, and one that an aggregate score cannot reveal.

\textbf{Contributions.}
(1) \textbf{\bench}, a negation benchmark for T2I models with $4{,}800$ prompts over $79$ object classes, six categories, and eight polarity settings, scored using only detection and geometry. Whereas prior negation evaluation for generation focuses on negated objects, \bench also negates attributes and relations while requiring the corresponding objects to remain present, and compares positive and negated prompts at matched prompt complexity.
(2) \textbf{An evaluator validation study} against human annotation on $600$ images, comparing \bench with 14 alternative scoring methods separately for positive and negated statements.
(3) \textbf{A fine-grained analysis of negation failures} across eleven T2I configurations, six categories, and eight polarity settings, including matched-load polarity comparisons and a statement-level failure taxonomy that reveals where and how models fail.

\section{Related Work}
\label{sec:related}
\textbf{Text-to-image models and compositional evaluation.}
Diffusion, flow-based, and autoregressive T2I models~\citep{ldm_2022rombach,sdxl_2023podell,sd3_2024esser,flux2_2025bfl,qwenimage_2025qwenteam,hidream_2025cai,zimage_2025team,hunyuanimage_2025tencent,januspro_2025chen,blip3onext_2025chen} and closed systems~\citep{dalle3_2023betker,imagen3_2024google,gptimage_2025openai} follow increasingly complex prompts. However, advances mostly show results for whether requested content appears, not whether a specified attribute or relation is withheld.
Distribution metrics such as FID~\citep{fid_2017heusel} and alignment scores such as CLIPScore~\citep{clipscore_2021hessel} give coarse quality and prompt following scores.
Like our approach, GenEval~\citep{geneval_2023ghosh} uses detectors to check objects, counts, colors, and positions, while PosEval~\citep{poseval_2025bader} extends its position task; \bench builds on GenEval's prompt schema and PosEval's spatial encoding. Other benchmarks cover more attributes, relations, and longer prompts~\citep{t2icompbench_2023huang,t2icompbenchpp_2025huang,dpgbench_2024hu,hrsbench_2023bakr,visor_2022gokhale,geneval2_2025kamath}.
GenAI-Bench~\citep{genaibench_2024li} and UniGenBench++~\citep{unigenbench_2025wang} include negation as a compositional skill or a grammar category, but neither separates attribute from relation negation or pairs negated prompts with positive ones. Detector-based benchmarks also face template contamination~\citep{genevalcontamination_2026xie}, weakening detector-human agreement as models improve~\citep{geneval2_2025kamath}, and sensitivity to wording~\citep{conceptmixpp_2025gan}; we therefore report human agreement by category and keep wording fixed across polarity conditions.

\textbf{Negation in text-to-image generation.}
T2I models still render what a prompt forbids, across several languages~\citep{logic_t2i_2024conwell,nosyndrome_2024nadeem}.
NegGenBench~\citep{vsf_2026guo} contains 200 prompt pairs and evaluates negative-prompt guidance: each prompt is paired with a separate negative prompt naming the content to exclude, instead of expressing the negation in the prompt itself.
LMD~\citep{lmd_2024lian} tests 100 scene-level prompts without a named object, scoring with an open-vocabulary detector and supplying the denied object via a negative-prompt input.
NegationCLIP~\citep{negationclip_2025park} evaluates 107 negated prompts with an MLLM judge, while NEG-TTOI~\citep{negtoi_2025cai} pairs a wanted object with an unwanted one in 2,000 prompts scored by GPT-4o. These studies establish that exclusion is hard, but none compares positive and negated versions of attributes and relations while requiring every named object to remain. A separate line of work steers generation away from unwanted content with negative prompts or guidance~\citep{cfg_2021ho,composablediffusion_2022liu,cptp_2026stirling,attendexcite_2023chefer,answer_2025desai}. While steering generation with negative prompts can exclude unwanted content, it does not test whether the model follows a negation written in the prompt, which is what \bench measures.

\textbf{Judging negation.}
Image-text and preference scores such as CLIPScore~\citep{clipscore_2021hessel}, BLIPScore~\citep{blip_2022li}, ImageReward~\citep{imagereward_2023xu}, PickScore~\citep{pickscore_2023kirstain}, and HPSv2~\citep{hpsv2_2023wu} measure overall alignment, but performance on ordinary prompts does not indicate negation reliability: CLIP-family models perform near chance on negated captions~\citep{negbench_2025alhamoud}, VQA models perform much worse on negated questions~\citep{negvqa_2025zhang}, and CC-Neg~\citep{conclip_2025singh} provides negated captions for training. VQA-based methods such as TIFA~\citep{tifa_2023hu}, which uses mPLUG~\citep{mplug_2022li}, and VQAScore~\citep{vqascore_2024lin} verify prompt requirements directly, but VQA judges, including Qwen3-VL~\citep{qwen3vl_2025qwen} and InternVL3.5~\citep{internvl35_2025wang}, can hallucinate absent details~\citep{dsg_2024cho,finer_2026xiao}, which matters when the evaluator must confirm that an attribute or relation is absent while the objects are present.
\citet{negationgeometry_2026sammani} propose a decomposed MLLM judge for negated image retrieval, but in our validation such judges lose up to 18.8 $\kappa$ points when negations are phrased as negations (Section~\ref{sec:evaluator-validation}). We therefore score each element with dedicated perception components and report human agreement by category and polarity.

\section{The \bench Benchmark}
\label{sec:benchmark}
\providecommand{\posstmt}[1]{\textcolor{blue}{#1}}
\providecommand{\negstmt}[1]{\textcolor{magenta}{#1}}
\begin{table}[t]
\caption{\textbf{\bench categories.} Each category has 800 prompts (100 prompts per polarity setting, 8 settings, 4,800 prompts over 79 object classes). Each row shows a matched pair from the benchmark: the same objects and value, differing only in polarity (\posstmt{positive}, \negstmt{negated}). Examples omit the shared prefix: ``a photo of'' for attributes, and ``a photo of a $\langle$subject$\rangle$ and a $\langle$target$\rangle$, where'' for relations.}
\label{tab:benchmark-overview}
\centering\small
\setlength{\tabcolsep}{4pt}
\begin{tabular}{@{}l p{0.30\linewidth} p{0.24\linewidth} p{0.24\linewidth}@{}}
\toprule
Category & Values & Positive example & Negated example \\
\midrule
\multicolumn{4}{@{}l}{\textit{Attributes}} \\
Color (10) & red, orange, yellow, green, blue, purple, pink, brown, black, white & a \posstmt{red} cup & a \negstmt{non-red} cup \\
Material (4) & wooden, metal, plastic, glass & a \posstmt{glass} cup & a \negstmt{non-glass} cup \\
\midrule
\multicolumn{4}{@{}l}{\textit{Relations}} \\
Directional (4) & left of, right of, above, below & the backpack is \posstmt{above} the parking meter & the backpack is \negstmt{not above} the parking meter \\
Proximity (2) & near, far from & the backpack is \posstmt{far from} the parking meter & the backpack is \negstmt{not far from} the parking meter \\
Comparative (4) & bigger than, smaller than, taller than, shorter than & the backpack is \posstmt{bigger than} the parking meter & the backpack is \negstmt{not bigger than} the parking meter \\
Depth (2) & in front of, behind & the backpack is \posstmt{in front of} the parking meter & the backpack is \negstmt{not in front of} the parking meter \\
\bottomrule
\end{tabular}
\end{table}

\label{sec:task}
\label{sec:taxonomy}
\label{sec:challenging}

\textbf{Task formulation.}
\bench measures how models respond when attributes and relations are negated. It tests both the effect of negation at a fixed number of statements and the effect of combining multiple statements in one prompt. Each prompt comes with metadata listing required objects and statements. A statement's polarity specifies whether an attribute or relation must hold (positive) or must not hold (negated); all named objects remain required in either case. The eight settings vary the number of positive and negated statements; we write a setting as P:N, so 1:1 contains one of each, and call P+N its load. To capture diverse forms of negation in prompts, we cover two attribute and four relation categories (Table~\ref{tab:benchmark-overview}), illustrated by the examples in Figure~\ref{fig:teaser}. Relation prompts link one shared subject to a different target for each statement, so $M$ statements name $M+1$ objects and changing polarity leaves both counts fixed. To support reliable automatic evaluation, we use COCO~\citep{coco_2014lin} object classes recognized by the detector and attributes and relations that our pipeline can verify in both polarities.

\textbf{Benchmark construction.}
We use GenEval's color vocabulary and adapt the position rules~\citep{geneval_2023ghosh}, select material values from OVAD~\citep{ovad_2023bravo}, and follow PosEval's relation encoding~\citep{poseval_2025bader}. An LLM~\citep{qwen3_2025qwenteam} selects combinations of objects, attributes, and relations from fixed lists. It keeps combinations that are plausible or unusual but possible to depict, so each negated statement denies something that could reasonably appear. Rule-based templates turn these selections into prompts with consistent wording. We balance different types of objects, attributes, and relations by favoring the ones that are used least often. The benchmark contains 4,800 prompts, with 100 for each combination of six categories and eight settings. Each category includes all 79 eligible object classes. Appendix~\ref{app:prompt-construction} gives the settings, selection rules, and templates; Appendices~\ref{app:evaluation-details} and~\ref{app:dataset-statistics} give the full vocabularies and distributions.

\subsection{Detector-based evaluation}
\label{sec:evaluation-protocol}
\begin{figure*}[t]
    \centering
    \includegraphics[width=\textwidth, trim={0 0 10.5cm 0}, clip]{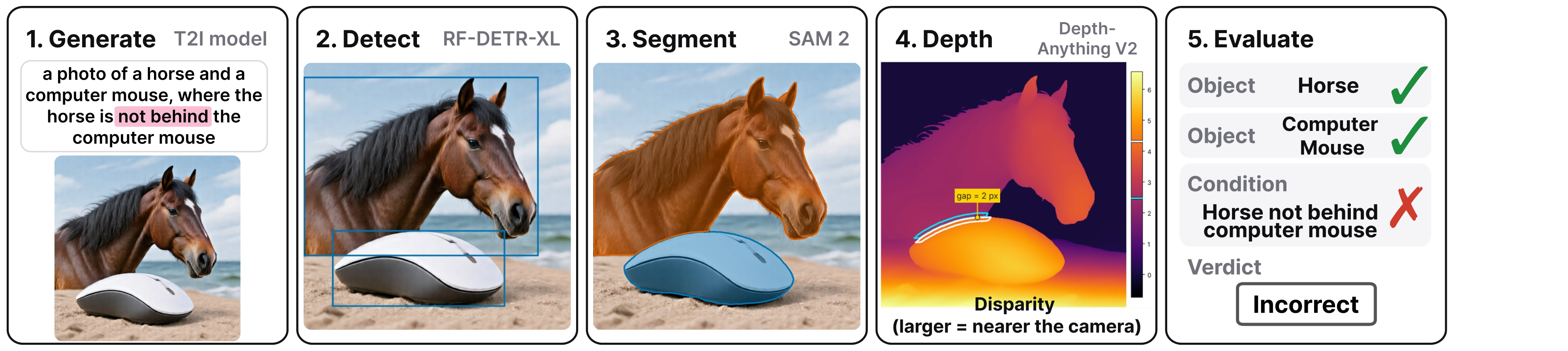}
    \caption{\textbf{\bench scores a negated depth relation.} RF-DETR-XL detects the objects, SAM~2 refines their masks, and Depth-Anything V2 estimates depth. Both objects are present, so the image cannot pass by omission, but the horse is behind the computer mouse and the negated statement fails. Checking each requirement separately lets the verdict name which one failed.}
    \label{fig:pipeline}
\end{figure*}

We check each prompt requirement separately to identify why an image fails. This could be a missing object, an incorrect count, or a violated attribute or relation. Each category uses the same verification rule (Appendix Table~\ref{tab:decision-rules}), making decisions reproducible and comparable across generators. Checking that every named object is present first prevents a model from satisfying a negated statement by omitting the object it refers to. Figure~\ref{fig:pipeline} illustrates this evaluation pipeline: both objects are present, but the horse is behind the computer mouse, violating the negated relation. We use RF-DETR-XL~\citep{rfdetr_2026robinson} to detect objects, SAM~2~\citep{sam2_2025ravi} to refine their masks, and Depth-Anything V2~\citep{depthanythingv2_2024yang} to estimate depth for front-behind relations.

\textbf{Statement decisions.}
For color and material, SigLIP 2~\citep{siglip2_2025tschannen} scores the masked object crop against the category's values. For color, a positive attribute passes if its score is within 0.5 of the highest score; a negated attribute passes if its score is more than 0.5 below it. This prevents a near-tied negated value from being treated as absent. For material, the top-scoring value decides both polarities. For relations, positive statements pass when their rule holds, and negated statements pass when it does not. Directional, comparative, and depth rules can return no relation when measurements are ambiguous (a dead zone).
Positive statements then fail, while the negated one passes because the denied relation was not found. Table~\ref{tab:decision-rules} defines the rules and final thresholds. Appendix~\ref{app:evaluation-details} gives the full equations and implementation details, and Appendix~\ref{app:deadzones} reports the frequency of ambiguous passes. Appendix~\ref{app:evaluator-ablations} compares detectors, mask refinement, and relation thresholds on a small subset of $140$ images with a clear human-majority verdict, from $600$ annotated images.

\textbf{Metrics. }
\textbf{Image accuracy} is the fraction of generated images that satisfy the full prompt: all required objects appear with valid counts, and every positive and negated statement passes. \textbf{Prompt accuracy} is the fraction of prompts for which at least one of four generated images is correct. The first measures success per image; the second measures success within four attempts.
We report image accuracy by category and setting, with overall scores averaged equally across the six categories; prompt accuracy is released with the per-prompt results. Each failed statement is also assigned one of five failure causes, which Section~\ref{sec:experiments} uses to diagnose where models fail; failure and metric definitions appear in Appendix~\ref{app:metric-definitions}.

\section{Validation of \bench's Evaluation}
\label{sec:evaluator-validation}
Since \bench scores images automatically, its verdicts matter only if they match human judgment. We test this with a human study and compare \bench with other evaluators.

\textbf{Human reference.}
A pool of 36 annotators labeled 600 generated images, with three labels per image, answering one yes/no question per object and per statement. We broke down each prompt into its objects and statements and phrased each statement as a question about whether it holds: for ``a non-wooden bench'', annotators say whether the bench is wooden, and the correct answer is ``no''. The majority vote on each element gives two references: a binary image label, true only if every element is correct, and a continuous score, the fraction of correct elements. Some images are ambiguous, so we also report an inter-annotator score (H. Inter-annotator), computed by scoring each annotator against the majority of the other two (Appendix~\ref{sec:sup-ceiling}). Evaluators are scored against the majority of all three annotators, a less noisy reference, so they can exceed this value.

\textbf{Evaluators.}
We compare \bench (Section~\ref{sec:evaluation-protocol}) with eight VQA judges: mPLUG~\citep{mplug_2022li}, VQAScore~\citep{vqascore_2024lin}, Qwen2.5-VL-7B~\citep{qwen25vl_2025bai}, InternVL3-8B~\citep{internvl3_zhu2025}, Qwen3-VL-8B and 32B~\citep{qwen3vl_2025qwen}, and InternVL3.5-8B and 38B~\citep{internvl35_2025wang}. Each judge answers yes/no questions in three modes: one question for the whole prompt (whole), the annotators' per-element questions (pos.), and the same questions with negated statements asked in negated form (neg.). The probability of the correct answer produces a continuous score, while an answer with a probability greater than $0.5$ yields a binary verdict. In per-element mode, an image passes only if every element receives the correct answer, similar to human labeling. \bench runs one detector pass per image, so it has a single mode. We also test six image-text similarity and reward scores, which have no binary form (see Appendix~\ref{sec:sup-measures}).

\begin{table*}[t]
\caption{\textbf{Cost and agreement with human labels} on 600 images, against the majority of three annotators. Judge modes: whole prompt (whole), per-element questions (pos), and per-element with negated statements asked in negated form (neg); $\Delta = \text{neg} - \text{pos}$. Precision and recall use the per-element mode. H.\ Inter-annotator: agreement between annotators. Bold: best in column. $^{\dagger}$Timed through t2v-metrics, which scores every image against every question, so time is overestimated.}

\label{tab:val-main}
\centering
\setlength{\tabcolsep}{3pt}
\resizebox{\textwidth}{!}{\begin{tabular}{@{}l rrr cccr cc cccr@{}}
\toprule
& \multicolumn{3}{c}{Cost} & \multicolumn{4}{c}{Cohen's $\kappa$} & \multicolumn{2}{c}{Per element} & \multicolumn{4}{c}{Kendall $\tau$} \\
\cmidrule(lr){2-4} \cmidrule(lr){5-8} \cmidrule(lr){9-10} \cmidrule(lr){11-14}
Evaluator & Params\,$\downarrow$ & GiB\,$\downarrow$ & s/img\,$\downarrow$ & whole\,$\uparrow$ & pos\,$\uparrow$ & neg\,$\uparrow$ & $\Delta$ & Prec.\,$\uparrow$ & Rec.\,$\uparrow$ & whole\,$\uparrow$ & pos\,$\uparrow$ & neg\,$\uparrow$ & $\Delta$ \\
\midrule
InternVL3.5-38B & 38B & 73.0 & 3.29 & 48.4 & 69.5 & 52.6 & -16.9 & 81.0 & 78.6 & 49.2 & 59.7 & 52.8 & -6.9 \\
Qwen3-VL-32B & 32B & 62.8 & 0.99 & 48.2 & \textbf{74.3} & \textbf{66.5} & -7.8 & \textbf{84.1} & 82.0 & 48.6 & \textbf{63.3} & \textbf{58.8} & -4.5 \\
VQAScore$^{\dagger}$ & 11B & 24.2 & 3.40 & 33.6 & 49.9 & 48.6 & \textbf{-1.3} & 68.4 & 65.0 & 38.8 & 42.3 & 42.8 & \textbf{0.5} \\
Qwen3-VL-8B & 8B & 16.8 & 0.49 & 48.2 & 69.7 & 54.1 & -15.6 & 81.7 & 78.2 & 46.4 & 61.0 & 54.8 & -6.2 \\
InternVL3-8B & 8B & 15.8 & 0.71 & 25.2 & 59.7 & 41.8 & -17.9 & 74.7 & 71.8 & 46.2 & 56.3 & 51.1 & -5.2 \\
InternVL3.5-8B & 8B & 17.1 & 0.76 & 35.3 & 69.0 & 50.2 & -18.8 & 77.9 & 82.0 & 43.2 & 55.3 & 49.8 & -5.5 \\
Qwen2.5-VL-7B$^{\dagger}$ & 7B & 15.9 & 8.39 & 35.2 & 52.8 & 44.3 & -8.5 & 79.6 & 56.8 & 44.6 & 43.7 & 44.2 & \textbf{0.5} \\
mPLUG & \textbf{0.6B} & 6.5 & \textbf{0.32} & 7.9 & 16.0 & 10.1 & -5.9 & 67.9 & 17.5 & 21.1 & -4.3 & -5.5 & -1.2 \\
\bench (ours) & 1.24B & \textbf{6.3} & 0.78 & \textbf{72.0} & -- & -- & -- & 79.2 & \textbf{85.0} & \textbf{65.1} & -- & -- & -- \\
\midrule
\textcolor{gray}{H. Inter-annotator} & \textcolor{gray}{--} & \textcolor{gray}{--} & \textcolor{gray}{--} & \textcolor{gray}{65.6} & \textcolor{gray}{--} & \textcolor{gray}{--} & \textcolor{gray}{--} & \textcolor{gray}{--} & \textcolor{gray}{--} & \textcolor{gray}{71.1} & \textcolor{gray}{--} & \textcolor{gray}{--} & \textcolor{gray}{--} \\
\bottomrule
\end{tabular}}
\end{table*}

\textbf{Agreement with humans.}
Table~\ref{tab:val-main} reports Cohen's $\kappa$~\citep{cohenkappa_1960cohen}, precision and recall against the binary label (per-element mode for judges), and Kendall $\tau$~\citep{kendall_1938kendall} against the continuous score. \bench reaches $\kappa=72.0$, above the human inter-annotator score ($65.6$), and $\tau=65.1$, above the best judge, Qwen3-VL-32B ($63.3$; bootstrap intervals overlap, Table~\ref{tab:sup-continuous}).
On $\kappa$, only Qwen3-VL-32B scores higher ($74.3$), while the other seven judges score below \bench on both measures.
Size adds little: scaling Qwen3-VL from $8$B to $32$B gains $4.6$ points of $\kappa$, and InternVL3.5 from $8$B to $38$B gains $0.5$. Similarity and reward scores reach at most $\tau=27.7$ and cannot tell whether a negated requirement holds (Table~\ref{tab:sup-continuous}).
In Table~\ref{tab:val-main}, \bench has the highest recall of any evaluator ($85.0$ vs. $82.0$ for the best judge) and trails it on precision ($79.2$ vs.\ $84.1$), so it rejects fewer human-accepted images than any judge: the failures we report are not an artifact of a strict evaluator (Appendix~\ref{sec:sup-confusion}).

\textbf{Negative phrasing effect.}
When we ask negated statements in negated form (``is the bench non-wooden?'') and keep all other questions the same ($\Delta$ in Table~\ref{tab:val-main}),
all judges lose 1.3 to 18.8 points of $\kappa$, and five of them lose 4.5 to 6.9 points of $\tau$, including the largest models.
Judge verdicts thus depend on wording, the very variable a negation benchmark changes. \bench checks each requirement directly instead of answering a question, so wording does not affect it, and it beats every judge under whole prompt and negated phrasing (Appendix~\ref{sec:sup-negation}).

\textbf{Cost.}
The \bench detector-based evaluator uses $1.24$B parameters and scores an image in $0.78$\,s with $6.3$\,GiB of GPU memory, measured on one GPU with one process and no batching. Judges with similar agreement use $8$B to $38$B parameters, $2.7$ to $11.6$ times more memory ($16.8$ to $73.0$\,GiB), and $0.49$ - $3.29$\,s per image, measured per question and multiplied by the mean of $4.66$ questions per image. Their cost also grows with the number of elements, since they ask one question per element, while \bench runs its detector once per image (Appendix~\ref{sec:sup-cost}). Its components can also be replaced as better detectors appear (Appendix~\ref{app:evaluator-ablations}).

\section{Benchmarking text-to-image models}
\label{sec:experiments}

We evaluate eleven T2I models on the full \bench corpus: BLIP3-o NEXT SFT 3B~\citep{blip3onext_2025chen}, FLUX.2-klein-9B and FLUX.2-dev (4-bit)~\citep{flux2_2025bfl}, HiDream-I1 (4-bit)~\citep{hidream_2025cai}, HunyuanImage-2.1 (distilled)~\citep{hunyuanimage_2025tencent}, Janus-Pro-7B~\citep{januspro_2025chen}, Qwen-Image~\citep{qwenimage_2025qwenteam}, SD~3.5 Large~\citep{sd3_2024esser}, Z-Image-Turbo~\citep{zimage_2025team}, and SD~1.5~\citep{ldm_2022rombach} with and without negation-aware text encoder NegationCLIP~\citep{negationclip_2025park}. Each configuration produces four samples for 4,800 prompts: 19,200 images per configuration, totaling 211,200 (Appendix~\ref{app:experiments}).

Scoring uses Section~\ref{sec:evaluation-protocol}'s evaluator: RF-DETR-XL~\citep{rfdetr_2026robinson} at presence/counting thresholds $0.03/0.5$, SAM~2 masks~\citep{sam2_2025ravi} for relations, SigLIP 2~\citep{siglip2_2025tschannen} for color and material, and Depth-Anything V2~\citep{depthanythingv2_2024yang} for depth. Geometric cut points and attribute confusion buffer are fitted on 140 human-annotated images, then fixed for the whole benchmark (Appendix~\ref{app:evaluator-ablations}). Prompts are sampled four times. Each model uses its native configuration: 4 to 50 steps, and default resolution from $384^2$ to $1024^2$. Appendix~\ref{app:generation-settings} lists exact configurations.

\begin{table*}[t]
\caption{\textbf{Accuracy (\%) on \bench, by polarity and statement load (left) and by semantic category (right).} Even the strongest models struggle with negation, and accuracy falls sharply as load increases. $P{:}N$ denotes $P$ asserted and $N$ negated statements; category values are averaged over all eight settings. Bold marks the best value per column.
}
\label{tab:main-accuracy}
\label{tab:setting-accuracy}
\label{tab:category-accuracy}
\centering
\setlength{\tabcolsep}{3pt}
\resizebox{\textwidth}{!}{\begin{tabular}{@{}lcccccccc|cccccc|c@{}}
\toprule
& \multicolumn{8}{c}{Statement polarity and load}
& \multicolumn{6}{c}{Semantic category} & \\
\cmidrule(lr){2-9}\cmidrule(lr){10-15}
& \multicolumn{2}{c}{Load 1}
& \multicolumn{3}{c}{Load 2}
& \multicolumn{2}{c}{Load 3}
& Load 4
& \multicolumn{2}{c}{Attributes}
& \multicolumn{4}{c}{Relations} & \\
\cmidrule(lr){2-3}\cmidrule(lr){4-6}
\cmidrule(lr){7-8}\cmidrule(lr){9-9}
\cmidrule(lr){10-11}\cmidrule(lr){12-15}
Model & 1:0 & 0:1 & 2:0 & 1:1 & 0:2 & 2:1 & 1:2 & 2:2
& Color & Material & Compar. & Depth & Direct. & Proxim. & Overall \\
\midrule
\multicolumn{16}{@{}l}{\textit{U-Net diffusion}} \\
\addlinespace[2pt]
SD~1.5
& 36.1 & 44.1 & 7.3 & 9.7 & 14.0 & 1.8 & 2.4 & 0.5
& 27.7 & 22.8 & 8.9 & 8.8 & 9.6 & 9.1 & 14.5 \\
SD~1.5 + NegationCLIP
& 35.7 & 44.3 & 8.8 & 9.6 & 15.5 & 1.8 & 3.0 & 0.5
& 28.3 & 23.9 & 8.4 & 9.4 & 10.0 & 9.2 & 14.9 \\
\midrule
\multicolumn{16}{@{}l}{\textit{Diffusion / flow transformers}} \\
\addlinespace[2pt]
HiDream-I1 (4-bit)
& 54.8 & 43.9 & 27.7 & 17.2 & 24.1 & 11.0 & 9.2 & 4.1
& 31.2 & 28.4 & 18.9 & 21.7 & 23.9 & 20.0 & 24.0 \\
Z-Image-Turbo
& 63.7 & 31.4 & 41.5 & 19.9 & 14.3 & 12.4 & 9.0 & 4.9
& 24.9 & 25.8 & 21.4 & 23.0 & 28.6 & 24.0 & 24.6 \\
SD~3.5 Large
& 57.9 & 47.5 & 29.3 & 17.8 & 23.8 & 8.7 & 8.2 & 3.8
& 31.4 & 36.4 & 18.5 & 19.7 & 22.5 & 19.3 & 24.6 \\
Qwen-Image
& 67.5 & 35.2 & 38.3 & 21.0 & 19.1 & 11.9 & 10.1 & 5.2
& 24.2 & 28.1 & 24.3 & 25.5 & 30.3 & 23.8 & 26.0 \\
FLUX.2-klein-9B
& 72.8 & 29.0 & 49.0 & 20.0 & 14.6 & 12.9 & 9.8 & 4.8
& 24.8 & 36.8 & 23.7 & 21.5 & 29.2 & 23.8 & 26.6 \\
HunyuanImage-2.1 (distilled)
& 66.2 & 44.7 & 41.2 & 26.2 & 25.1 & \textbf{17.9} & 12.9 & \textbf{8.4}
& 29.3 & 36.7 & 25.2 & \textbf{27.6} & 36.6 & 26.6 & 30.3 \\
FLUX.2-dev (4-bit)
& \textbf{76.3} & 42.6 & \textbf{50.1} & 22.8 & 23.8 & 16.1 & 12.2 & 6.4
& 27.9 & \textbf{44.2} & \textbf{25.4} & 26.6 & 37.2 & 26.3 & 31.3 \\
\midrule
\multicolumn{16}{@{}l}{\textit{Autoregressive / hybrid}} \\
\addlinespace[2pt]
Janus-Pro-7B
& 55.2 & \textbf{54.8} & 31.0 & 21.8 & 25.8 & 9.7 & 9.9 & 4.7
& \textbf{39.8} & 38.0 & 19.5 & 20.6 & 22.4 & 19.3 & 26.6 \\
BLIP3-o NEXT SFT 3B
& 67.2 & 48.7 & 41.7 & \textbf{27.8} & \textbf{28.1} & 16.2 & \textbf{15.0} & 6.2
& 34.7 & 37.1 & 24.5 & 26.6 & \textbf{38.1} & \textbf{27.1} & \textbf{31.4} \\
\bottomrule
\end{tabular}}
\end{table*}

Table~\ref{tab:main-accuracy} reports accuracy by polarity, statement load, and semantic category.
The strongest model, BLIP3-o NEXT, reaches $31.4\%$.
Scores vary by about $\pm0.7$ points overall and $\pm2$ points per subtask by chance alone (95\% interval). The top three models lie within 1.1 points, so overall score cannot separate them, but subtasks can.
Janus-Pro-7B ties FLUX.2-klein-9B fourth at $26.6\%$ overall, yet first on negated single statement at $54.8\%$, ahead of BLIP3-o NEXT at $48.7\%$, and first on color at $39.8\%$. FLUX.2-dev (4-bit) is the opposite: first on the asserted single statement at $76.3\%$, and $42.6\%$ once negated. Accuracy on one negated statement spans $29.0$ to $54.8\%$, a wider range than overall score. Four models lead at least one subtask, and none leads more than three of eight polarity subtasks or two of six category subtasks.

\textbf{Polarity and compositional load.}
\label{sec:matched-load}
At Load 1, nine of eleven configurations score higher on the asserted statement (1:0) than on the negated one (0:1); the exceptions are the two SD~1.5 variants. At Load 2, the same nine score highest on the all-positive setting (2:0). Across those nine, drop averages $22.7$ points, from $0.4$ to $43.8$.
Load matters on its own. Between 1:1 and 0:2 there is no consistent ordering, so after the first negation the number of statements matters as much as how many are negated. Accuracy then collapses as load grows: the best score in each group falls from $76.3\%$ at Load 1 to $17.9\%$ at Load 3 and $8.4\%$ at Load 4. \bench exposes two separate difficulties: negation is harder than assertion at equal load, and satisfying several constraints at once is hard whatever their polarity.

Penalty depends on category. At Load 1, the drop is largest for the attributes: $53.7$ points for color, from $85.0\%$ asserted to $31.3\%$ denied, and $17.0$ points for material. Relations lose less, $16.6$ points for comparative, $14.6$ for directional. Proximity barely moves, $45.8\%$ vs. $43.2\%$. Depth is opposite, $41.7\%$ vs. $44.0\%$.
This reversal is largely explained by dead-zone passes rather than by better negation
Categories with higher positive accuracy also lose more when negated (Spearman $\rho = 0.94$ across the six categories). A larger drop therefore does not by itself show that a category is harder to negate; it may partly reflect a higher starting point (see Appendix~\ref{app:additional-results} for more details).

Two caveats apply when reading these comparisons. First, the size of a model's drop does not rank its negation ability, because the drop is capped by the model's accuracy on the positive statement: FLUX.2-klein-9B scores $72.8\%$ at 1:0 and loses $43.8$ points at 0:1, while Janus-Pro-7B scores $55.2\%$, loses $0.4$, and still reaches the highest negated accuracy. A model that is weak on the assertion has less to lose on the denial, so only the within-model comparison is safe to read.
Second, \bench measures whether a negated attribute or relation is absent from the image, not whether the model understood the negation. A negated statement is easier to satisfy by chance than a positive one, so a model that ignores the negation can still pass some negated statements.
A control prompt that omits the tested attribute or relation would separate the two; it fits the current design and is left for future work (Appendix~\ref{app:limitations}).

\begin{figure*}[t]
\centering
\begin{minipage}[t]{0.49\textwidth}
\centering
\includegraphics[width=\linewidth]{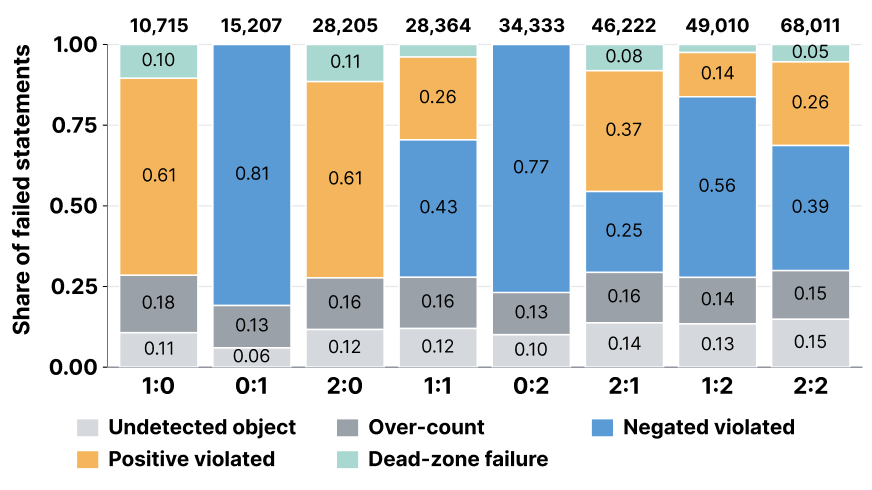}
\textbf{(a) By benchmark setting}
\end{minipage}
\hfill
\begin{minipage}[t]{0.49\textwidth}
\centering
\includegraphics[width=\linewidth]{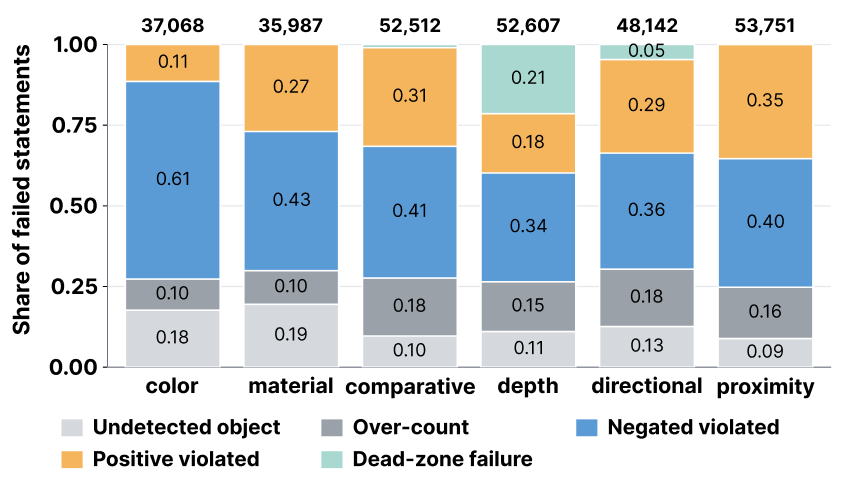}
\textbf{(b) By category}
\end{minipage}
\caption{\textbf{Statement-level failure decomposition, pooled over all models.} Each bar partitions the failed statements of one polarity setting (a) or category (b) into the five failure causes; the number above each bar is its count of failed statements. Bars show how failures divide among causes, not how often statements fail. Each failed statement is counted once; statements that were satisfied are not counted, even in failed images. Dead zones only affect positive statements in directional, comparative, and depth relations, so the other bars have no dead-zone failures.}
\label{fig:statement-failures}
\end{figure*}

\textbf{Failure analysis.}
\label{sec:failures}
Every failed statement falls into one of five causes: undetected object, over-count, positive violation, negated violation, or dead-zone failure, defined in Appendix~\ref{app:metric-definitions}. Of the 280,067 failed statements, negated violations are the largest group at $41.5\%$. Positive violations follow at $25.9\%$, then over-counts at $15.0\%$, undetected objects at $12.6\%$, and dead-zone failures at $5.0\%$. The most common failure is a model rendering what the prompt forbade. Dead zones, the only cause attributable to the evaluator alone, are the smallest at $5.0\%$.
Within the same prompt, negated statements fail more often than positive ones (Figure~\ref{fig:statement-failures}(a)). At 1:1, the negated statement causes 43\% of failures and the positive one 26\%; at 1:2, the two negated statements cause 56\% and the positive one 14\%, twice as much per statement.
Per statement, the imbalance holds in all four mixed settings, by a factor of $1.3$ to $2.0$.
The share of negated violations also varies by category (Figure~\ref{fig:statement-failures}(b)): they account for 61\% of failures for color and 43\% for material, but 34 to 41\% for the four relations. Negated colors are thus most often rendered, resulting in an incorrect image.
Polarity therefore changes which constraint a model breaks, not only how often it breaks one. Appendix~\ref{app:additional-results} gives the per-model and row-level breakdowns.

\textbf{Dead zones.}
\label{sec:deadzones}
Directional, comparative, and depth relations have deliberate dead zones. When the measured geometry is ambiguous, the evaluator returns no predicate rather than forcing one. This is often the correct outcome: two objects at similar depth are neither \emph{in front of} nor \emph{behind} each other. Proximity has no dead zone, because \emph{near} and \emph{far} are exact complements.
Dead zones affect the two polarities differently. An asserted statement fails when its predicate is absent. A negated statement passes when the forbidden predicate is absent, including when no predicate is resolved at all. Such a pass is valid under the benchmark rule, but rests on weaker evidence than one where an alternative relation is established.
Of the negated relation statements that pass, 19.0\% pass only because no relation was resolved. This happens mostly for depth (55.5\%, vs.\ 9.2\% for directional and 2.9\% for comparative), which explains why depth is the only category where negated statements score higher than positive ones at Load 1.
Table~\ref{tab:deadzone-exposure} in Appendix~\ref{app:deadzones} reports exposure per category.

\textbf{Negation-aware text encoders.}
\label{sec:negationclip}
NegationCLIP~\citep{negationclip_2025park} replaces SD~1.5's text encoder and raises overall accuracy from $14.5\%$ to $14.9\%$. Both SD~1.5 variants score higher at 0:1 than at 1:0, and both stay far below the other models. The comparison is not clean: the two runs share a checkpoint and a 50-step schedule, but use guidance scales of 9.0 and 7.5. It is suggestive rather than a causal encoder test (Appendix~\ref{app:additional-results}).
Representing ``not red'' differently from ``red'' helps only if the generator can act on the distinction. The same applies to a classifier-free-guidance negative prompt, which steers sampling away from unwanted content without the model interpreting the negation~\citep{cfg_2021ho,negationclip_2025park,negtoi_2025cai}. Improving negation is therefore likely to require training or conditioning on negation-bearing supervision, not an encoder swap or a negative-prompt field. \bench measures whether such methods work, both on attributes and relations.

\begin{figure*}[t]
    \centering
    \includegraphics[width=\textwidth]{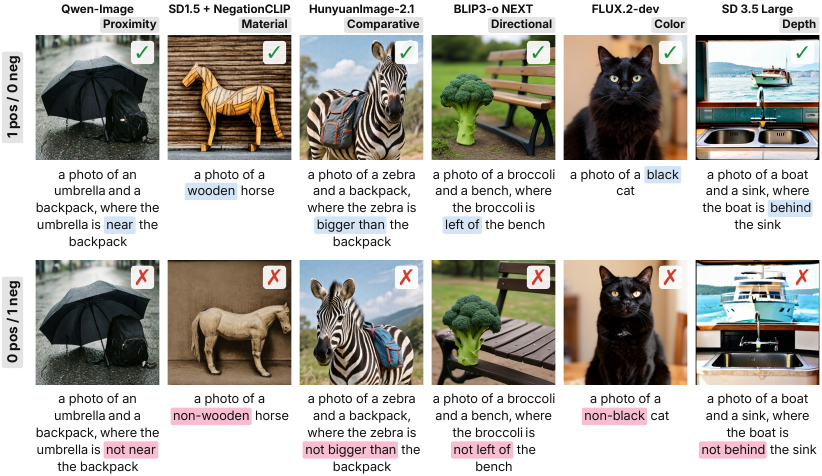}
    \caption{\textbf{Qualitative comparison of positive and negated prompts.}
    Each column shows one model and a different semantic category:
    proximity, material, comparative, directional, color, and depth.
    The top row contains single positive statements ($1{:}0$);
    the bottom row negates the corresponding statements ($0{:}1$),
    keeping the objects and tested predicates unchanged.
    Marks give the \bench verdict (\ding{51} pass, \ding{55} fail).
    In all six pairs, the model satisfies the positive prompt and fails the negated one.
    }
    \label{fig:qualitative-examples}
\end{figure*}

\textbf{Qualitative examples.}
\label{app:qualitative-examples}
Figure~\ref{fig:qualitative-examples} illustrates the distinction \bench is designed to measure: rendering a requested property versus excluding a forbidden one while preserving the named objects. Each column pairs a positive and a negated prompt from one of six categories, generated by the same model. By changing polarity at fixed statement and object counts, these examples isolate the prompt change of interest. The selected failures show why successful positive-prompt generation alone does not establish negation-following ability, motivating our separate evaluation of positive and negated requirements.

\section{Conclusion}
\label{sec:conclusion}
We introduced \bench, a benchmark that measures whether T2I models can withhold what a prompt forbids while still rendering every object it names. It compares positive and negated prompts at matched complexity and object counts, guards against success by omission, and reduces accidental success. Its detector-based evaluator agrees with human labels comparably to much larger vision-language judges while using a fraction of their GPU memory, and, unlike these judges, does not depend on how the evaluation question is phrased.
Across eleven T2I configurations and 211,200 generated images, nine of eleven score lower on a single negated statement than on a single positive one, with a loss that depends on the category and compounds as statements are combined. Polarity also changes which model leads and which error it makes. Generating what a prompt asks for and withholding what it forbids are thus distinct abilities, and an aggregate compositional score hides the second. \bench measures it directly, providing a controlled testbed for future models and for methods that aim to improve negation.

\section*{Acknowledgements}

This work was partially funded by the ERC (853489-DEXIM) and the Alfried Krupp von Bohlen und Halbach Foundation, which we thank for their generous support.

The authors gratefully acknowledge the scientific support and resources of the AI service infrastructure \textit{LRZ AI Systems} provided by the Leibniz Supercomputing Centre (LRZ) of the Bavarian Academy of Sciences and Humanities (BAdW), funded by Bayerisches Staatsministerium f\"ur Wissenschaft und Kunst (StMWK).

The authors gratefully acknowledge the Gauss Centre for Supercomputing e.V. (\href{http://www.gauss-centre.eu}{www.gauss-centre.eu}) for funding this project by providing computing time on the GCS Supercomputer JUWELS at J\"ulich Supercomputing Centre (JSC).

\bibliography{references}

@inproceedings{geneval_2023ghosh,
  title     = {{GenEval}: An Object-Focused Framework for Evaluating Text-to-Image Alignment},
  author    = {Ghosh, Dhruba and Hajishirzi, Hannaneh and Schmidt, Ludwig},
  booktitle = {NeurIPS},
  year      = {2023}
}

@inproceedings{coco_2014lin,
  title     = {Microsoft {COCO}: Common Objects in Context},
  author    = {Lin, Tsung-Yi and Maire, Michael and Belongie, Serge and Hays, James and Perona, Pietro and Ramanan, Deva and Doll{\'a}r, Piotr and Zitnick, C. Lawrence},
  booktitle = {ECCV},
  year      = {2014}
}

@inproceedings{ovad_2023bravo,
  title     = {Open-Vocabulary Attribute Detection},
  author    = {Bravo, Mar{\'i}a A. and Mittal, Sudhanshu and Ging, Simon and Brox, Thomas},
  booktitle = {CVPR},
  year      = {2023}
}

@inproceedings{maskrcnn_2017he,
  title     = {Mask {R-CNN}},
  author    = {He, Kaiming and Gkioxari, Georgia and Doll{\'a}r, Piotr and Girshick, Ross},
  booktitle = {ICCV},
  year      = {2017}
}

@article{siglip2_2025tschannen,
  title   = {{SigLIP 2}: Multilingual Vision-Language Encoders with Improved Semantic Understanding, Localization, and Dense Features},
  author={Tschannen, Michael and Gritsenko, Alexey and Wang, Xiao and Naeem, Muhammad Ferjad and Alabdulmohsin, Ibrahim and Parthasarathy, Nikhil and Evans, Talfan and Beyer, Lucas and Xia, Ye and Mustafa, Basil and H\'enaff, Olivier and Harmsen, Jeremiah and Steiner, Andreas and Zhai, Xiaohua},
  journal = {arXiv preprint arXiv:2502.14786},
  year    = {2025}
}

@inproceedings{depthanythingv2_2024yang,
  title     = {Depth Anything {V2}},
  author    = {Yang, Lihe and Kang, Bingyi and Huang, Zilong and Zhao, Zhen and Xu, Xiaogang and Feng, Jiashi and Zhao, Hengshuang},
  booktitle = {NeurIPS},
  year      = {2024}
}

@article{qwen3_2025qwenteam,
  title   = {Qwen3. Technical Report},
  author  = {Yang, An and Li, Anfeng and Yang, Baosong and Zhang, Beichen and Hui, Binyuan and Zheng, Bo and Yu, Bowen and Gao, Chang and Huang, Chengen and Lv, Chenxu and Zheng, Chujie and Liu, Dayiheng and Zhou, Fan and Huang, Fei and Hu, Feng and Ge, Hao and Wei, Haoran and Lin, Huan and Tang, Jialong and Yang, Jian and Tu, Jianhong and Zhang, Jianwei and Yang, Jianxin and Yang, Jiaxi and Zhou, Jing and Zhou, Jingren and Lin, Junyang and Dang, Kai and Bao, Keqin and Yang, Kexin and Yu, Le and Deng, Lianghao and Li, Mei and Xue, Mingfeng and Li, Mingze and Zhang, Pei and Wang, Peng and Zhu, Qin and Men, Rui and Gao, Ruize and Liu, Shixuan and Luo, Shuang and Li, Tianhao and Tang, Tianyi and Yin, Wenbiao and Ren, Xingzhang and Wang, Xinyu and Zhang, Xinyu and Ren, Xuancheng and Fan, Yang and Su, Yang and Zhang, Yichang and Zhang, Yinger and Wan, Yu and Liu, Yuqiong and Wang, Zekun and Cui, Zeyu and Zhang, Zhenru and Zhou, Zhipeng and Qiu, Zihan},  
  journal = {arXiv preprint arXiv:2505.09388},
  year    = {2025}
}

@inproceedings{ldm_2022rombach,
  title     = {High-Resolution Image Synthesis with Latent Diffusion Models},
  author    = {Rombach, Robin and Blattmann, Andreas and Lorenz, Dominik and Esser, Patrick and Ommer, Bj{\"o}rn},
  booktitle = {CVPR},
  year      = {2022}
}

@inproceedings{sdxl_2023podell,
  title     = {{SDXL}: Improving Latent Diffusion Models for High-Resolution Image Synthesis},
  author    = {Podell, Dustin and English, Zion and Lacey, Kyle and Blattmann, Andreas and Dockhorn, Tim and M{\"u}ller, Jonas and Penna, Joe and Rombach, Robin},
  booktitle = {ICLR},
  year      = {2024}
}

@inproceedings{sd3_2024esser,
  title     = {Scaling Rectified Flow Transformers for High-Resolution Image Synthesis},
  author    = {Esser, Patrick and Kulal, Sumith and Blattmann, Andreas and Entezari, Rahim and M{\"u}ller, Jonas and Saini, Harry and Levi, Yam and Lorenz, Dominik and Sauer, Axel and Boesel, Frederic and Podell, Dustin and Dockhorn, Tim and English, Zion and Lacey, Kyle and Goodwin, Alex and Marek, Yannik and Rombach, Robin},
  booktitle = {ICML},
  year      = {2024}
}

@misc{flux2_2025bfl,
  title   = {{FLUX.2} black-forest-labs/FLUX.2-klein-9B},
  author  = {{Black Forest Labs}},
  url     = {https://blackforestlabs.ai},
  year    = {2026}
}

@article{qwenimage_2025qwenteam,
  title   = {{Qwen-Image} Technical Report},
  author  = {Wu, Chenfei and Li, Jiahao and Zhou, Jingren and Lin, Junyang and Gao, Kaiyuan and Yan, Kun and Yin, Sheng-ming and Bai, Shuai and Xu, Xiao and Chen, Yilei and Chen, Yuxiang and Tang, Zecheng and Zhang, Zekai and Wang, Zhengyi and Yang, An and Yu, Bowen and Cheng, Chen and Liu, Dayiheng and Li, Deqing and Zhang, Hang and Meng, Hao and Wei, Hu and Ni, Jingyuan and Chen, Kai and Cao, Kuan and Peng, Liang and Qu, Lin and Wu, Minggang and Wang, Peng and Yu, Shuting and Wen, Tingkun and Feng, Wensen and Xu, Xiaoxiao and Wang, Yi and Zhang, Yichang and Zhu, Yongqiang and Wu, Yujia and Cai, Yuxuan and Liu, Zenan},
  journal = {arXiv preprint arXiv:2508.02324},
  year    = {2025}
}

@inproceedings{negationclip_2025park,
  title     = {Know ``No'' Better: A Data-Driven Approach for Enhancing Negation Awareness in {CLIP}},
  author    = {Park, Junsung and Lee, Jungbeom and Song, Jongyoon and Yu, Sangwon and Jung, Dahuin and Yoon, Sungroh},
  booktitle = {ICCV},
  year      = {2025}
}

@techreport{gptimage_2025openai,
  title   = {Addendum to {GPT-4o} System Card: Native Image Generation},
  author  = {{OpenAI}},
  institution = {OpenAI},
  year    = {2025},
  url     = {https://cdn.openai.com/11998be9-5319-4302-bfbf-1167e093f1fb/Native_Image_Generation_System_Card.pdf}
}

@misc{dalle3_2023betker,
  title  = {Improving Image Generation with Better Captions},
  author = {Betker, James and Goh, Gabriel and Jing, Li and Brooks, Tim and Wang, Jianfeng and Li, Linjie and Ouyang, Long and Zhuang, Juntang and Lee, Joyce and Guo, Yufei and others},
  url    = {https://cdn.openai.com/papers/dall-e-3.pdf},
  year   = {2023}
}

@article{imagen3_2024google,
  title   = {Imagen 3},
  author  = {Baldridge, Jason and Bauer, Jakob and Bhutani, Mukul and Brichtova, Nicole and Bunner, Andrew and Castrejon, Lluis and Chan, Kelvin and Chen, Yichang and Dieleman, Sander and Du, Yuqing and others},
  journal = {arXiv preprint arXiv:2408.07009},
  year    = {2024}
}

@inproceedings{cfg_2021ho,
  title     = {Classifier-Free Diffusion Guidance},
  author    = {Ho, Jonathan and Salimans, Tim},
  booktitle = {NeurIPS Workshop on Deep Generative Models and Downstream Applications},
  year      = {2021}
}

@inproceedings{composablediffusion_2022liu,
  title     = {Compositional Visual Generation with Composable Diffusion Models},
  author    = {Liu, Nan and Li, Shuang and Du, Yilun and Torralba, Antonio and Tenenbaum, Joshua B.},
  booktitle = {ECCV},
  year      = {2022}
}

@article{attendexcite_2023chefer,
  title   = {Attend-and-Excite: Attention-Based Semantic Guidance for Text-to-Image Diffusion Models},
  author  = {Chefer, Hila and Alaluf, Yuval and Vinker, Yael and Wolf, Lior and Cohen-Or, Daniel},
  journal = {ACM Transactions on Graphics},
  year    = {2023}
}

@inproceedings{t2icompbench_2023huang,
  title     = {{T2I-CompBench}: A Comprehensive Benchmark for Open-world Compositional Text-to-Image Generation},
  author    = {Huang, Kaiyi and Sun, Kaiyue and Xie, Enze and Li, Zhenguo and Liu, Xihui},
  booktitle = {NeurIPS},
  year      = {2023}
}

@article{poseval_2025bader,
  title   = {Stitch: Training-Free Position Control in Multimodal Diffusion Transformers},
  author  = {Bader, Jessica and Pach, Mateusz and Bravo, Mar{\'i}a A. and Belongie, Serge and Akata, Zeynep},
  journal = {arXiv preprint arXiv:2509.26644},
  year    = {2025}
}

@article{dpgbench_2024hu,
  title   = {{ELLA}: Equip Diffusion Models with {LLM} for Enhanced Semantic Alignment},
  author  = {Hu, Xiwei and Wang, Rui and Fang, Yixiao and Fu, Bin and Cheng, Pei and Yu, Gang},
  journal = {arXiv preprint arXiv:2403.05135},
  year    = {2024}
}

@inproceedings{hrsbench_2023bakr,
  title     = {{HRS-Bench}: Holistic, Reliable and Scalable Benchmark for Text-to-Image Models},
  author    = {Bakr, Eslam Mohamed and Sun, Pengzhan and Shen, Xiaoqian and Khan, Faizan Farooq and Li, Li Erran and Elhoseiny, Mohamed},
  booktitle = {ICCV},
  year      = {2023}
}

@article{visor_2022gokhale,
  title   = {Benchmarking Spatial Relationships in Text-to-Image Generation},
  author  = {Gokhale, Tejas and Palangi, Hamid and Nushi, Besmira and Vineet, Vibhav and Horvitz, Eric and Kamar, Ece and Baral, Chitta and Yang, Yezhou},
  journal = {arXiv preprint arXiv:2212.10015},
  year    = {2022}
}

@inproceedings{fid_2017heusel,
  title     = {{GANs} Trained by a Two Time-Scale Update Rule Converge to a Local Nash Equilibrium},
  author    = {Heusel, Martin and Ramsauer, Hubert and Unterthiner, Thomas and Nessler, Bernhard and Hochreiter, Sepp},
  booktitle = {NeurIPS},
  year      = {2017}
}

@inproceedings{clipscore_2021hessel,
  title     = {{CLIPScore}: A Reference-Free Evaluation Metric for Image Captioning},
  author    = {Hessel, Jack and Holtzman, Ari and Forbes, Maxwell and Le Bras, Ronan and Choi, Yejin},
  booktitle = {EMNLP},
  year      = {2021}
}

@inproceedings{tifa_2023hu,
  title     = {{TIFA}: Accurate and Interpretable Text-to-Image Faithfulness Evaluation with Question Answering},
  author    = {Hu, Yushi and Liu, Benlin and Kasai, Jungo and Wang, Yizhong and Ostendorf, Mari and Krishna, Ranjay and Smith, Noah A.},
  booktitle = {ICCV},
  year      = {2023}
}

@inproceedings{vqascore_2024lin,
  title     = {Evaluating Text-to-Visual Generation with Image-to-Text Generation},
  author    = {Lin, Zhiqiu and Pathak, Deepak and Li, Baiqi and Li, Jiayao and Xia, Xide and Neubig, Graham and Zhang, Pengchuan and Ramanan, Deva},
  booktitle = {ECCV},
  year      = {2024}
}

@inproceedings{dsg_2024cho,
  title     = {Davidsonian Scene Graph: Improving Reliability in Fine-Grained Evaluation for Text-to-Image Generation},
  author    = {Cho, Jaemin and Hu, Yushi and Garg, Roopal and Anderson, Peter and Krishna, Ranjay and Baldridge, Jason and Bansal, Mohit and Pont-Tuset, Jordi and Wang, Su},
  booktitle = {ICLR},
  year      = {2024}
}

@inproceedings{negbench_2025alhamoud,
  title     = {Vision-Language Models Do Not Understand Negation},
  author    = {Alhamoud, Kumail and Alshammari, Shaden and Tian, Yonglong and Li, Guohao and Torr, Philip H. S. and Kim, Yoon and Ghassemi, Marzyeh},
  booktitle = {CVPR},
  year      = {2025}
}

@inproceedings{finer_2026xiao,
  title     = {{FINER}: {MLLMs} Hallucinate under Fine-Grained Negative Queries},
  author    = {Xiao, Rui and Kim, Sanghwan and Xian, Yongqin and Akata, Zeynep and Alaniz, Stephan},
  booktitle = {CVPR},
  year      = {2026}
}

@inproceedings{genevalcontamination_2026xie,
  title     = {Reconstruction Alignment Improves Unified Multimodal Models},
  author    = {Xie, Ji and Darrell, Trevor and Zettlemoyer, Luke and Wang, XuDong},
  booktitle = {ICLR},
  year      = {2026}
}

@inproceedings{pope_2023li,
  title     = {Evaluating Object Hallucination in Large Vision-Language Models},
  author    = {Li, Yifan and Du, Yifan and Zhou, Kun and Wang, Jinpeng and Zhao, Wayne Xin and Wen, Ji-Rong},
  booktitle = {EMNLP},
  year      = {2023}
}

@article{januspro_2025chen,
  title   = {{Janus-Pro}: Unified Multimodal Understanding and Generation with Data and Model Scaling},
  author  = {Chen, Xiaokang and Wu, Zhiyu and Liu, Xingchao and Pan, Zizheng and Liu, Wen and Xie, Zhenda and Yu, Xingkai and Ruan, Chong},
  journal = {arXiv preprint arXiv:2501.17811},
  year    = {2025}
}

@inproceedings{hidream_2025cai,
  title     = {{HiDream-I1}: An Open-Source High-Efficient Image Generative Foundation Model},
  author    = {Cai, Qi and Chen, Jingwen and Chen, Yang and Li, Yehao and Long, Fuchen and Pan, Yingwei and Qiu, Zhaofan and Zhang, Yiheng and Gao, Fengbin and Xu, Peihan and others},
  booktitle = {ACM International Conference on Multimedia},
  year      = {2025}
}

@article{zimage_2025team,
  title   = {{Z-Image}: An Efficient Image Generation Foundation Model with Single-Stream Diffusion Transformer},
  author  = {Cai, Huanqia and Cao, Sihan and Du, Ruoyi and Gao, Peng and Hoi, Steven and Hou, Zhaohui and Huang, Shijie and Jiang, Dengyang and Jin, Xin and Li, Liangchen and Li, Zhen and Li, Zhong-Yu and Liu, David and Liu, Dongyang and Shi, Junhan and Wu, Qilong and Yu, Feng and Zhang, Chi and Zhang, Shifeng and Zhou, Shilin},
  journal = {arXiv preprint arXiv:2511.22699},
  year    = {2025}
}

@inproceedings{rfdetr_2026robinson,
  title     = {{RF-DETR}: Neural Architecture Search for Real-Time Detection Transformers},
  author    = {Robinson, Isaac and Robicheaux, Peter and Popov, Matvei and Ramanan, Deva and Peri, Neehar},
  booktitle = {ICLR},
  year      = {2026}
}

@inproceedings{sam2_2025ravi,
  title     = {{SAM} 2: Segment Anything in Images and Videos},
  author    = {Ravi, Nikhila and Gabeur, Valentin and Hu, Yuan-Ting and Hu, Ronghang and Ryali, Chaitanya and Ma, Tengyu and Khedr, Haitham and R{\"a}dle, Roman and Rolland, Chloe and Gustafson, Laura and Mintun, Eric and Pan, Junting and Alwala, Kalyan Vasudev and Carion, Nicolas and Wu, Chao-Yuan and Girshick, Ross and Doll{\'a}r, Piotr and Feichtenhofer, Christoph},
  booktitle = {ICLR},
  year      = {2025}
}

@article{t2icompbenchpp_2025huang,
  title   = {{T2I-CompBench++}: An Enhanced and Comprehensive Benchmark for Compositional Text-to-Image Generation},
  author  = {Huang, Kaiyi and Duan, Chengqi and Sun, Kaiyue and Xie, Enze and Li, Zhenguo and Liu, Xihui},
  journal = {TPAMI},
  year    = {2025}
}

@article{genaibench_2024li,
  title={Genai-bench: Evaluating and improving compositional text-to-visual generation},
  author={Li, Baiqi and Lin, Zhiqiu and Pathak, Deepak and Li, Jiayao and Fei, Yixin and Wu, Kewen and Ling, Tiffany and Xia, Xide and Zhang, Pengchuan and Neubig, Graham and others},
  journal={arXiv preprint arXiv:2406.13743},
  year={2024}
}

@article{unigenbench_2025wang,
  title   = {{UniGenBench++}: A Unified Semantic Evaluation Benchmark for Text-to-Image Generation},
  author  = {Wang, Yibin and Li, Zhimin and Zang, Yuhang and Bu, Jiazi and Zhou, Yujie and Xin, Yi and He, Junjun and Wang, Chunyu and Lu, Qinglin and Jin, Cheng and Wang, Jiaqi},
  journal = {arXiv preprint arXiv:2510.18701},
  year    = {2025}
}

@article{geneval2_2025kamath,
  title   = {{GenEval} 2: Addressing Benchmark Drift in Text-to-Image Evaluation},
  author  = {Kamath, Amita and Chang, Kai-Wei and Krishna, Ranjay and Zettlemoyer, Luke and Hu, Yushi and Ghazvininejad, Marjan},
  journal = {arXiv preprint arXiv:2512.16853},
  year    = {2025}
}

@article{negtoi_2025cai,
  title   = {{TNG-CLIP}: Training-Time Negation Data Generation for Negation Awareness of {CLIP}},
  author  = {Cai, Yuliang and Thomason, Jesse and Rostami, Mohammad},
  journal = {arXiv preprint arXiv:2505.18434},
  year    = {2025}
}

@article{lmd_2024lian,
  title   = {{LLM}-Grounded Diffusion: Enhancing Prompt Understanding of Text-to-Image Diffusion Models with Large Language Models},
  author  = {Lian, Long and Li, Boyi and Yala, Adam and Darrell, Trevor},
  journal = {TMLR},
  year    = {2024}
}

@inproceedings{vsf_2026guo,
  title     = {{VSF}: Simple, Efficient, and Effective Negative Guidance in Few-Step Image Generation Models by Value Sign Flip},
  author    = {Guo, Wenqi and Du, Shan},
  booktitle = {ICLR},
  year      = {2026}
}

@inproceedings{negvqa_2025zhang,
  title     = {{NegVQA}: Can Vision Language Models Understand Negation?},
  author    = {Zhang, Yuhui and Su, Yuchang and Liu, Yiming and Yeung-Levy, Serena},
  booktitle = {Findings of ACL},
  year      = {2025}
}

@article{logic_t2i_2024conwell,
  title   = {Relations, Negations, and Numbers: Looking for Logic in Generative Text-to-Image Models},
  author  = {Conwell, Colin and Tawiah-Quashie, Rupert and Ullman, Tomer},
  journal = {arXiv preprint arXiv:2411.17066},
  year    = {2024}
}

@inproceedings{conclip_2025singh,
  title     = {Learning the Power of ``No'': Foundation Models with Negations},
  author    = {Singh, Jaisidh and Shrivastava, Ishaan and Vatsa, Mayank and Singh, Richa and Bharati, Aparna},
  booktitle = {WACV},
  year      = {2025}
}

@inproceedings{negationgeometry_2026sammani,
  title     = {When Negation Is a Geometry Problem in Vision-Language Models},
  author    = {Sammani, Fawaz and Chamiti, Tzoulio and Gavrikov, Paul and Deligiannis, Nikos},
  booktitle = {CVPR Workshop on Multimodal Algorithmic Reasoning},
  year      = {2026}
}

@article{nosyndrome_2024nadeem,
  title   = {Negation Blindness in Large Language Models: Unveiling the {NO} Syndrome in Image Generation},
  author  = {Nadeem, Mohammad and Sohail, Shahab Saquib and Cambria, Erik and Schuller, Bj{\"o}rn W. and Hussain, Amir},
  journal = {arXiv preprint arXiv:2409.00105},
  year    = {2024}
}

@article{conceptmixpp_2025gan,
  title   = {{ConceptMix++}: Leveling the Playing Field in Text-to-Image Benchmarking via Iterative Prompt Optimization},
  author  = {Gan, Haosheng and Tinaz, Berk and Sepehri, Mohammad Shahab and Fabian, Zalan and Soltanolkotabi, Mahdi},
  journal = {arXiv preprint arXiv:2507.03275},
  year    = {2025}
}

@inproceedings{answer_2025desai,
  title     = {Guiding Diffusion Models with Adaptive Negative Sampling Without External Resources},
  author    = {Desai, Alakh and Vasconcelos, Nuno},
  booktitle = {ICCV},
  year      = {2025}
}

@inproceedings{cptp_2026stirling,
  title     = {Controllable Image Generation with Composed Parallel Token Prediction},
  author    = {Stirling, Jamie and Al-Moubayed, Noura and Willcocks, Chris G. and Shum, Hubert P. H.},
  booktitle = {CVPR Workshop on Low-Level Vision Frontiers},
  year      = {2026}
}

@article{blip3onext_2025chen,
  title   = {{BLIP3o-NEXT}: Next Frontier of Native Image Generation},
  author  = {Chen, Jiuhai and Xue, Le and Xu, Zhiyang and Pan, Xichen and Yang, Shusheng
             and Qin, Can and Yan, An and Zhou, Honglu and Chen, Zeyuan
             and Huang, Lifu and others},
  journal = {arXiv preprint arXiv:2510.15857},
  year    = {2025},
}

@misc{hunyuanimage_2025tencent,
  title={HunyuanImage 2.1: An Efficient Diffusion Model for High-Resolution (2K) Text-to-Image Generation},
  author={{Tencent Hunyuan Team}},
  year={2025},
  howpublished={\url{https://github.com/Tencent-Hunyuan/HunyuanImage-2.1}},
}

@inproceedings{imagereward_2023xu,
  title     = {{ImageReward}: Learning and Evaluating Human Preferences for Text-to-Image Generation},
  author    = {Xu, Jiazheng and Liu, Xiao and Wu, Yuchen and Tong, Yuxuan and Li, Qinkai and Ding, Ming and Tang, Jie and Dong, Yuxiao},
  booktitle = {NeurIPS},
  year      = {2023}
}

@inproceedings{pickscore_2023kirstain,
  title     = {Pick-a-Pic: An Open Dataset of User Preferences for Text-to-Image Generation},
  author    = {Kirstain, Yuval and Polyak, Adam and Singer, Uriel and Matiana, Shahbuland and Penna, Joe and Levy, Omer},
  booktitle = {NeurIPS},
  year      = {2023}
}

@article{hpsv2_2023wu,
  title   = {Human Preference Score v2: A Solid Benchmark for Evaluating Human Preferences of Text-to-Image Synthesis},
  author  = {Wu, Xiaoshi and Hao, Yiming and Sun, Keqiang and Chen, Yixiong and Zhu, Feng and Zhao, Rui and Li, Hongsheng},
  journal = {arXiv preprint arXiv:2306.09341},
  year    = {2023}
}

@inproceedings{mplug_2022li,
  title     = {{mPLUG}: Effective and Efficient Vision-Language Learning by Cross-modal Skip-connections},
  author    = {Li, Chenliang and Xu, Haiyang and Tian, Junfeng and Wang, Wei and Yan, Ming and Bi, Bin and Ye, Jiabo and Chen, Hehong and Xu, Guohai and Cao, Zheng and Zhang, Ji and Huang, Songfang and Huang, Fei and Zhou, Jingren and Si, Luo},
  booktitle = {EMNLP},
  year      = {2022}
}

@article{qwen25vl_2025bai,
  title   = {{Qwen2.5-VL} Technical Report},
  author  = {Bai, Shuai and Chen, Keqin and Liu, Xuejing and Wang, Jialin and Ge, Wenbin and Song, Sibo and Dang, Kai and Wang, Peng and Wang, Shijie and Tang, Jun and others},
  journal = {arXiv preprint arXiv:2502.13923},
  year    = {2025}
}

@article{internvl3_zhu2025,
  title={Internvl3: Exploring advanced training and test-time recipes for open-source multimodal models},
  author={Zhu, Jinguo and Wang, Weiyun and Chen, Zhe and Liu, Zhaoyang and Ye, Shenglong and Gu, Lixin and Tian, Hao and Duan, Yuchen and Su, Weijie and Shao, Jie and others},
  journal={arXiv preprint arXiv:2504.10479},
  year={2025}
}

@article{qwen3vl_2025qwen,
  title   = {{Qwen3-VL} Technical Report},
  author  = {Bai, Shuai and Cai, Yuxuan and Chen, Ruizhe and Chen, Keqin and Chen, Xionghui and Cheng, Zesen and Deng, Lianghao and Ding, Wei and Gao, Chang and Ge, Chunjiang and Ge, Wenbin and Guo, Zhifang and Huang, Qidong and Huang, Jie and Huang, Fei and Hui, Binyuan and Jiang, Shutong and Li, Zhaohai and Li, Mingsheng and Li, Mei and Li, Kaixin and Lin, Zicheng and Lin, Junyang and Liu, Xuejing and Liu, Jiawei and Liu, Chenglong and Liu, Yang and Liu, Dayiheng and Liu, Shixuan and Lu, Dunjie and Luo, Ruilin and Lv, Chenxu and Men, Rui and Meng, Lingchen and Ren, Xuancheng and Ren, Xingzhang and Song, Sibo and Sun, Yuchong and Tang, Jun and Tu, Jianhong and Wan, Jianqiang and Wang, Peng and Wang, Pengfei and Wang, Qiuyue and Wang, Yuxuan and Xie, Tianbao and Xu, Yiheng and Xu, Haiyang and Xu, Jin and Yang, Zhibo and Yang, Mingkun and Yang, Jianxin and Yang, An and Yu, Bowen and Zhang, Fei and Zhang, Hang and Zhang, Xi and Zheng, Bo and Zhong, Humen and Zhou, Jingren and Zhou, Fan and Zhou, Jing and Zhu, Yuanzhi and Zhu, Ke},
  journal = {arXiv preprint arXiv:2511.21631},
  year    = {2025}
}

@article{internvl35_2025wang,
  title   = {{InternVL3.5}: Advancing Open-Source Multimodal Models in Versatility, Reasoning, and Efficiency},
  author  = {Wang, Weiyun and Gao, Zhangwei and Gu, Lixin and Pu, Hengjun and Cui, Long and Wei, Xingguang and Liu, Zhaoyang and Jing, Linglin and Ye, Shenglong and Shao, Jie and Wang, Zhaokai and Chen, Zhe and Zhang, Hongjie and Yang, Ganlin and Wang, Haomin and Wei, Qi and Yin, Jinhui and Li, Wenhao and Cui, Erfei and Chen, Guanzhou and Ding, Zichen and Tian, Changyao and Wu, Zhenyu and Xie, Jingjing and Li, Zehao and Yang, Bowen and Duan, Yuchen and Wang, Xuehui and Hou, Zhi and Hao, Haoran and Zhang, Tianyi and Li, Songze and Zhao, Xiangyu and Duan, Haodong and Deng, Nianchen and Fu, Bin and He, Yinan and Wang, Yi and He, Conghui and Shi, Botian and He, Junjun and Xiong, Yingtong and Lv, Han and Wu, Lijun and Shao, Wenqi and Zhang, Kaipeng and Deng, Huipeng and Qi, Biqing and Ge, Jiaye and Guo, Qipeng and Zhang, Wenwei and Zhang, Songyang and Cao, Maosong and Lin, Junyao and Tang, Kexian and Gao, Jianfei and Huang, Haian and Gu, Yuzhe and Lyu, Chengqi and Tang, Huanze and Wang, Rui and Lv, Haijun and Ouyang, Wanli and Wang, Limin and Dou, Min and Zhu, Xizhou and Lu, Tong and Lin, Dahua and Dai, Jifeng and Su, Weijie and Zhou, Bowen and Chen, Kai and Qiao, Yu and Wang, Wenhai and Luo, Gen},
  journal = {arXiv preprint arXiv:2508.18265},
  year    = {2025}
}

@inproceedings{blip_2022li,
  title     = {{BLIP}: Bootstrapping Language-Image Pre-training for Unified Vision-Language Understanding and Generation},
  author    = {Li, Junnan and Li, Dongxu and Xiong, Caiming and Hoi, Steven},
  booktitle = {ICML},
  year      = {2022}
}

@article{cohenkappa_1960cohen,
  title   = {A Coefficient of Agreement for Nominal Scales},
  author  = {Cohen, Jacob},
  journal = {Educational and Psychological Measurement},
  volume  = {20},
  number  = {1},
  pages   = {37--46},
  year    = {1960}
}

@article{kendall_1938kendall,
  title   = {A New Measure of Rank Correlation},
  author  = {Kendall, Maurice G.},
  journal = {Biometrika},
  volume  = {30},
  number  = {1--2},
  pages   = {81--93},
  year    = {1938}
}

@article{Shabani2022HouseDiffusionVF,
  title={HouseDiffusion: Vector Floorplan Generation via a Diffusion Model with Discrete and Continuous Denoising},
  author={Mohammad Amin Shabani and Sepidehsadat Hosseini and Yasutaka Furukawa},
  journal={CVPR},
  year={2023},
}

@article{Ko2022LargescaleTG,
  title={Large-scale Text-to-Image Generation Models for Visual Artists’ Creative Works},
  author={Hyung-Kwon Ko and Gwanmo Park and Hyeon Jeon and Jaemin Jo and Juho Kim and Jinwook Seo},
  journal={International Conference on Intelligent User Interfaces},
  year={2022},
}
\bibliographystyle{unsrtnat}

\appendix
\section{Benchmark construction}
\label{app:benchmark}

\subsection{Prompt construction and representative examples}
\label{app:prompt-construction}

\textbf{Polarity settings and object-count controls.}
The eight settings are \texttt{pos1\_neg0}, \texttt{pos2\_neg0}, \texttt{pos0\_neg1}, \texttt{pos0\_neg2}, \texttt{pos1\_neg1}, \texttt{pos1\_neg2}, \texttt{pos2\_neg1}, and \texttt{pos2\_neg2}. A prompt contains $M=P+N$ statements, each assigning an attribute to an object or specifying a relation between two objects. At a fixed statement load, polarity varies while the number of statements remains constant: for example, \texttt{pos2\_neg0}, \texttt{pos1\_neg1}, and \texttt{pos0\_neg2} each contain two statements. Attribute prompts name one object per statement. Relation prompts use a star topology with one shared subject $A$ and distinct targets $B_1,\ldots,B_M$, giving exactly $M+1$ named objects. Thus, within each category, comparisons at equal statement load also preserve object count. Negating an attribute or relation never removes its objects: ``a non-wooden bench'' still requires a bench, and ``a bicycle that is not left of a car'' still requires both objects.

\textbf{Taxonomy scope.}
The object vocabulary follows the detector-compatible COCO classes; prompts exclude \texttt{person}, leaving 79 eligible classes. Size is treated as a comparative relation because it is defined relative to another object, rather than as an intrinsic attribute.

\textbf{Templates and candidate selection.} Prompts are generated by deterministic templates (examples in Table~\ref{tab:prompt-examples}). Attribute prompts name each object with its value: ``a photo of a $\langle$value$\rangle$ $\langle$object$\rangle$'' for a positive statement and ``a photo of a non-$\langle$value$\rangle$ $\langle$object$\rangle$'' for a negated one. Relation prompts use the star topology: ``a photo of a $\langle$subject$\rangle$ and a $\langle$target$_1\rangle$ \dots, where the $\langle$subject$\rangle$ is [not] $\langle$relation$\rangle$ the $\langle$target$_i\rangle$'', with one clause per statement.

The LLM selector, \texttt{Qwen/Qwen3-30B-A3B-Instruct-2507}, filters object/value/predicate combinations from fixed vocabularies to those judged plausible or unusual but depictable in their positive form. This ensures that the denied attribute or relation is a meaningful possibility for the selected objects. Prompt strings are generated deterministically after selection, so wording does not drift across polarity conditions.
Negation is then written into the prompt itself, and the selector plays no role in the wording or in the evaluation.

For relation construction, the selector receives a shared subject, a candidate
pool of target objects, and the allowed predicates for that category. It returns
only allowed predicate-target pairs. Attribute construction analogously selects
values from the fixed color/material vocabularies. Returned values are validated
before entering the corpus, and selector outputs are cached and reused.
Among valid combinations, balancing takes priority over selector preference.
Relation subjects and targets have separate usage counters; candidate targets
are drawn from the least-used portion of the vocabulary and under-used
predicates are preferred. Attribute objects and values follow the analogous
policy.

\begin{table}[t]
\caption{\textbf{Representative prompts from the released corpus.}}
\label{tab:prompt-examples}
\centering
\begin{tabular}{p{2.15cm}p{1.55cm}p{8.0cm}}
\toprule
Category & Setting & Prompt \\
\midrule
Color &
\texttt{pos0\_neg1} &
a photo of a non-brown orange \\

Material &
\texttt{pos2\_neg2} &
a photo of a plastic bus, a plastic knife, a non-plastic sandwich, and a non-plastic toilet \\

Comparative &
\texttt{pos0\_neg1} &
a photo of a skis and a skateboard, where the skis is not bigger than the skateboard \\

Depth &
\texttt{pos0\_neg1} &
a photo of a broccoli and a bench, where the broccoli is not in front of the bench \\

Directional &
\texttt{pos2\_neg2} &
a photo of a laptop, a zebra, a backpack, a surfboard, and a carrot, where the laptop is right of the zebra, is right of the backpack, is not below the surfboard, and is not right of the carrot \\

Proximity &
\texttt{pos0\_neg1} &
a photo of a broccoli and a bench, where the broccoli is not far from the bench \\
\bottomrule
\end{tabular}
\end{table}

\textbf{Category vocabularies and sources.}
Table~\ref{tab:benchmark-overview} lists the complete vocabularies; every value or relation can be positive or negated. The color list is retained from GenEval~\citep{geneval_2023ghosh}. The four material values are selected from the broader OVAD vocabulary~\citep{ovad_2023bravo}. Structured prompts follow GenEval's metadata schema and PosEval's relation encoding~\citep{poseval_2025bader}. The category-specific evaluation rules are defined in Table~\ref{tab:decision-rules} and Appendix~\ref{app:evaluation-details}.

\subsection{Dataset statistics and balancing}
\label{app:dataset-statistics}

The final star-topology corpus contains 4,800 prompts: six categories, eight settings, and 100 prompts in every category-setting cell. Each category has 800 prompts and covers all 79 eligible COCO object classes. Averaged over the eight settings, a prompt contains 2.25 statements, half positive and half negated. Attribute prompts name one object per statement, giving 2.25 objects per prompt; relation prompts name one shared subject plus one target per statement, giving 3.25 objects per prompt.

\begin{figure*}[t]
    \centering
    \includegraphics[width=0.98\textwidth]{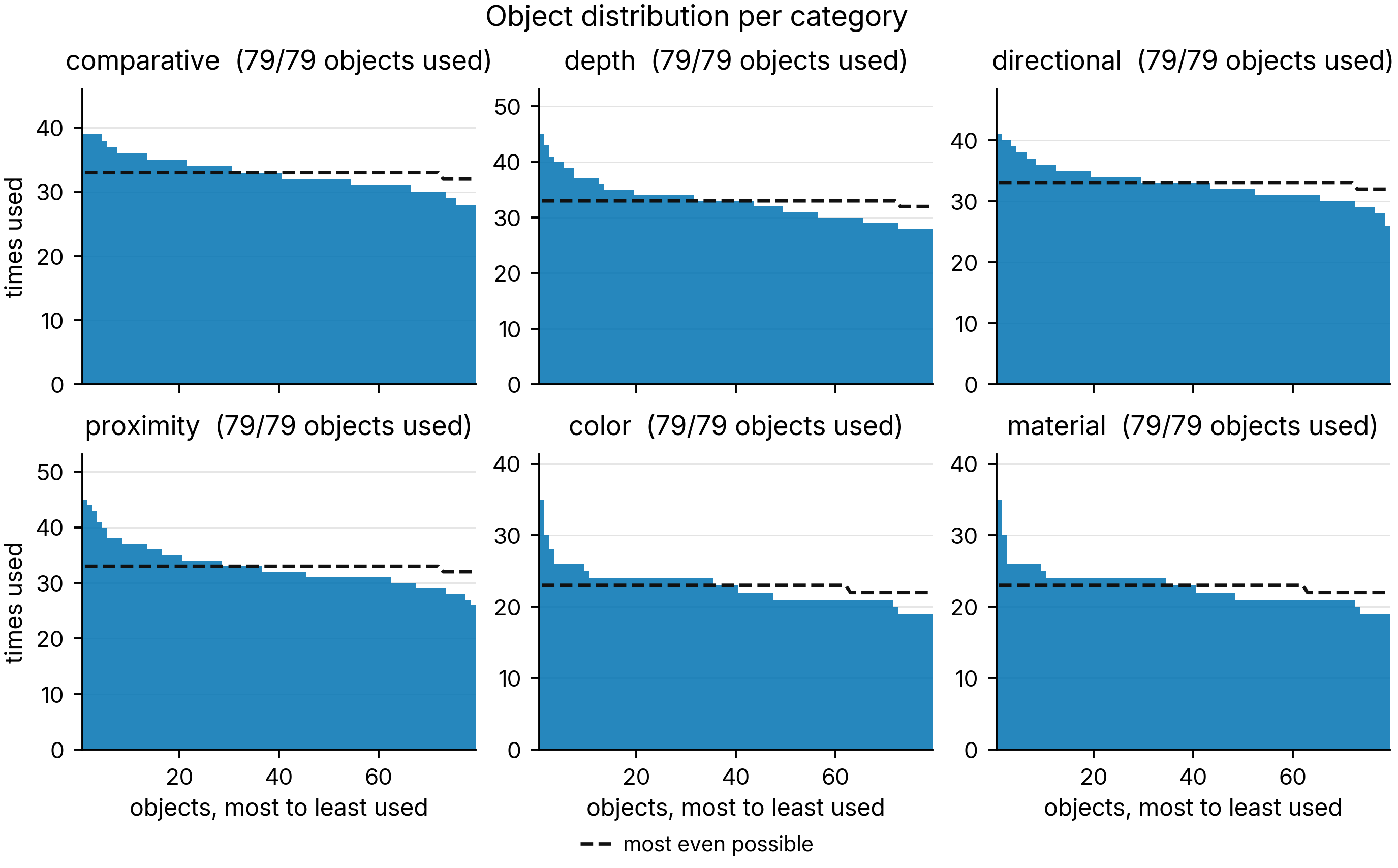}
    \caption{\textbf{Object distribution per category.}
    Each bar is one object's occurrence count, sorted from most to least used;
    the dashed line shows the most even allocation reachable for the observed
    number of draws. All 79 eligible COCO classes appear in every category.}
    \label{fig:object-frequency}
\end{figure*}

\begin{figure}[t]
    \centering
    \includegraphics[width=\linewidth]{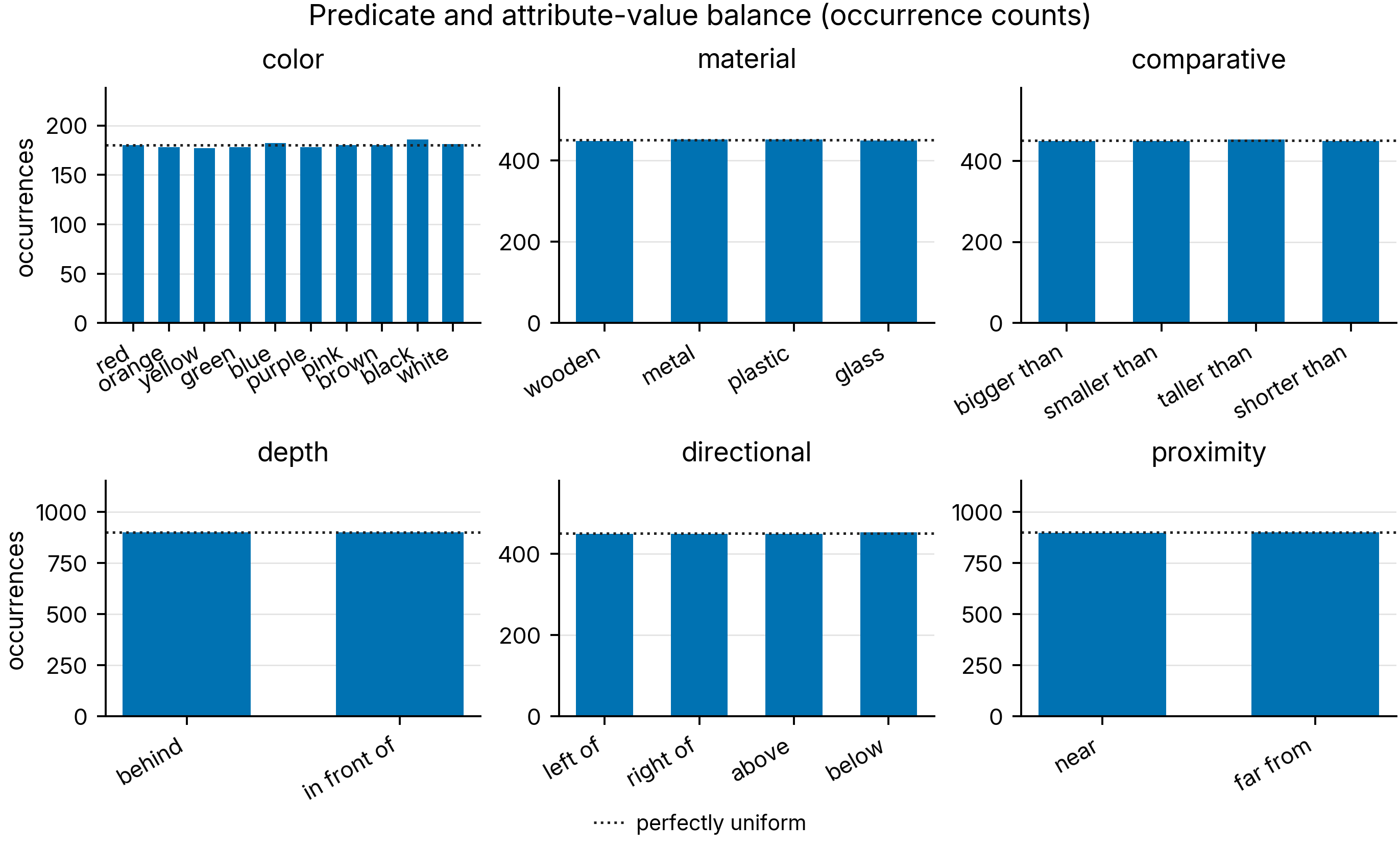}
    \caption{\textbf{Predicate and attribute-value balance.}
    Frequency of each relation predicate or attribute value within its category.}
    \label{fig:predicate-balance}
\end{figure}

As shown in Figures~\ref{fig:object-frequency} and~\ref{fig:predicate-balance}, all 79 eligible COCO object classes appear in every category. Pooled over the
eight settings, individual object counts remain close to the most-even
allocation achievable for the corresponding number of draws, and the smaller
predicate/value vocabularies are distributed approximately uniformly.

\section{Human study and evaluator validation}
\label{app:human-validation}

\begin{figure}[h!]
\centering
\includegraphics[width=0.82\textwidth, trim={0 0 0 2.8cm}, clip]{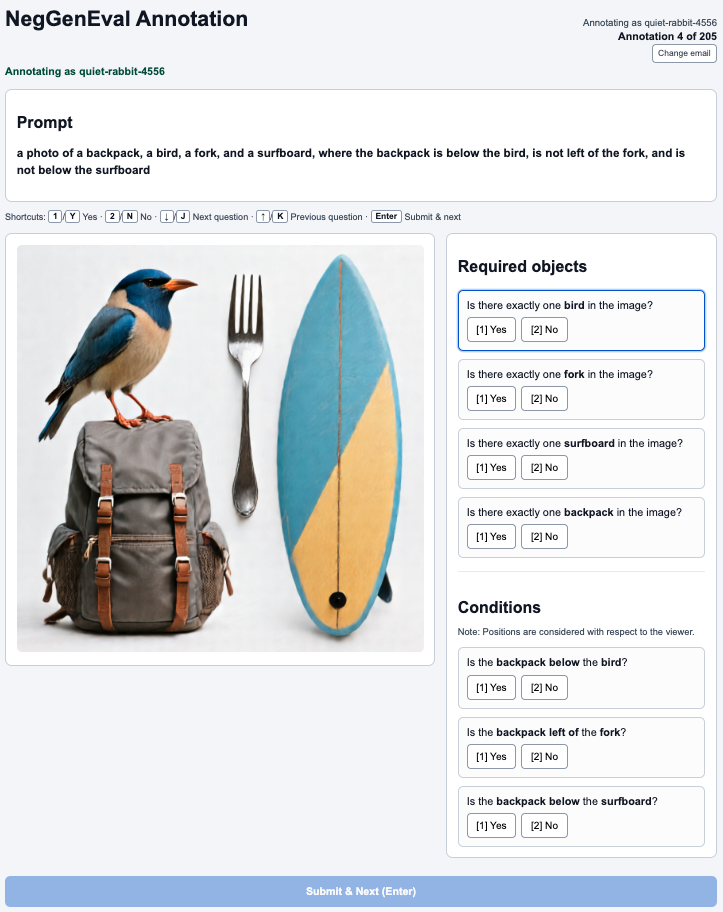}
\caption{\textbf{The annotation interface.} The prompt is shown in full, then one question per required object and one per statement. The prompt here negates two relations, but both are asked affirmatively: the annotator answers whether the backpack is left of the fork, and a correct image gives ``no''. Annotators see neither the expected answers nor the T2I generator.}
\label{fig:annotator-view}
\end{figure}

\begin{table}[t]
\centering
\caption{\textbf{Agreement between annotators.} Fleiss' $\kappa$ and the share of statements all three answered alike. Negated statements are harder for people as well, so we compare evaluators against the leave-one-out inter-annotator score rather than against perfect agreement.}
\label{tab:sup-annotators}
\begin{tabular}{lrrrr}
\toprule
Statements & $n$ & Fleiss' $\kappa$ & Unanimous & Leave-one-out $\kappa$ \\
\midrule
All statements & 2772 & 72.4 & 84.9 & 70.9 \\
Affirmative & 2249 & 73.0 & 88.3 & 73.3 \\
Negated & 523 & 60.2 & 70.2 & 59.7 \\
\bottomrule
\end{tabular}
\end{table}

\subsection{Reference labels and human inter-annotator score}
\label{sec:sup-ceiling}

In this study, 36 annotators labeled 600 images, three per image, using the interface in Figure~\ref{fig:annotator-view}. The reference for an element is the majority of the three answers. The binary image label is true only when every element of the image is satisfied, and the continuous image score is the fraction of satisfied elements.
We estimate the inter-annotator score (H. Inter-annotator) with a leave-one-out scheme: each annotator's labels are scored against the majority of the other two, and the score is averaged over the three held-out annotators.
Table~\ref{tab:sup-annotators} reports agreement between annotators: Fleiss' $\kappa$ is $72.4$ over all statements, and $84.9\%$ of statements are answered unanimously. Agreement is lower on negated statements ($\kappa=60.2$, $70.2\%$ unanimous) than on affirmative ones ($73.0$, $88.3\%$), so negation is harder for people as well.

The inter-annotator score is a conservative reference, not an upper bound on achievable agreement. Each held-out annotator is compared with the majority of two annotators, a noisier target than the three-annotator majority used for every evaluator. An evaluator above $\kappa=65.6$ therefore agrees with the three-annotator consensus more closely than a single annotator agrees with the other two; this does not make it more reliable than a human.

\subsection{Evaluators' scores}
\label{sec:sup-evaluators}
\begin{table}[t]
\centering
\caption{\textbf{Binary agreement, all scoring modes.} Whole prompt, the benchmark's per-element decomposition, and that decomposition with the statements the prompt negates asked in negated form; all other questions are unchanged.}
\label{tab:sup-binary-all}
\begin{tabular}{lrrr}
\toprule
Evaluator & Whole prompt & Per element & Negated per element \\
\midrule
\multicolumn{4}{l}{\emph{Detector-based}} \\
\bench (ours) & -- & \textbf{72.0} & \textbf{72.0} \\
\midrule
\multicolumn{4}{l}{\emph{VQA judges}} \\
VQAScore & 33.6 & 49.9 & 48.6 \\
mPLUG & 7.9 & 16.0 & 10.1 \\
Qwen2.5-VL-7B & 35.2 & 52.8 & 44.3 \\
InternVL3-8B & 25.2 & 59.7 & 41.8 \\
Qwen3-VL-8B & 48.2 & 69.7 & 54.1 \\
Qwen3-VL-32B & 48.2 & 74.3 & 66.5 \\
InternVL3.5-8B & 35.3 & 69.0 & 50.2 \\
InternVL3.5-38B & 48.4 & 69.5 & 52.6 \\
\midrule
H. Inter-annotator & -- & 65.6 & -- \\
\bottomrule
\end{tabular}
\end{table}

\begin{table}[t!]
\centering
\small
\caption{\textbf{Rank agreement with the human score.} Spearman $\rho$, Kendall $\tau$ and pairwise ranking accuracy, with 95\% bootstrap intervals over images. Judges are ordered by their best $\rho$. Rank agreement needs no threshold, so it is the only comparison that admits the similarity and reward scores.}
\label{tab:sup-continuous}
\begin{tabular}{lccc}
\toprule
Evaluator & $\rho$ & $\tau$ & Pair acc. \\
\midrule
\multicolumn{4}{l}{\emph{Qwen3-VL-32B~\citep{qwen3vl_2025qwen}}} \\
\quad whole prompt & 63.2\,\tiny{[57.6, 68.2]} & 48.6\,\tiny{[43.9, 52.9]} & 87.5\,\tiny{[81.7, 92.4]} \\
\quad pos. per element & \textbf{76.5}\,\tiny{[71.5, 81.1]} & 63.3\,\tiny{[58.7, 67.7]} & \textbf{94.7}\,\tiny{[91.6, 97.4]} \\
\quad neg. per element & 72.4\,\tiny{[67.0, 77.3]} & 58.8\,\tiny{[54.2, 63.3]} & 94.4\,\tiny{[91.3, 97.1]} \\
\midrule
\multicolumn{4}{l}{\emph{Qwen3-VL-8B~\citep{qwen3vl_2025qwen}}} \\
\quad whole prompt & 60.4\,\tiny{[54.2, 66.0]} & 46.4\,\tiny{[41.2, 51.1]} & 86.5\,\tiny{[81.4, 91.2]} \\
\quad pos. per element & 74.0\,\tiny{[68.5, 78.8]} & 61.0\,\tiny{[56.3, 65.6]} & 94.2\,\tiny{[91.2, 96.7]} \\
\quad neg. per element & 68.3\,\tiny{[62.6, 73.4]} & 54.8\,\tiny{[50.0, 59.5]} & 94.3\,\tiny{[91.2, 96.8]} \\
\midrule
\multicolumn{4}{l}{\emph{InternVL3.5-38B~\citep{internvl35_2025wang}}} \\
\quad whole prompt & 63.6\,\tiny{[57.9, 68.8]} & 49.2\,\tiny{[44.4, 53.7]} & 87.4\,\tiny{[81.7, 92.2]} \\
\quad pos. per element & 73.9\,\tiny{[68.7, 78.4]} & 59.7\,\tiny{[55.0, 64.0]} & 94.3\,\tiny{[91.4, 96.7]} \\
\quad neg. per element & 67.2\,\tiny{[61.5, 72.4]} & 52.8\,\tiny{[47.8, 57.5]} & 93.9\,\tiny{[91.1, 96.5]} \\
\midrule
\multicolumn{4}{l}{\emph{InternVL3-8B~\citep{internvl3_zhu2025}}} \\
\quad whole prompt & 60.1\,\tiny{[54.1, 65.6]} & 46.2\,\tiny{[41.2, 51.1]} & 86.1\,\tiny{[80.6, 91.0]} \\
\quad pos. per element & 69.5\,\tiny{[63.5, 74.6]} & 56.3\,\tiny{[51.2, 61.0]} & 92.3\,\tiny{[88.7, 95.4]} \\
\quad neg. per element & 64.9\,\tiny{[58.8, 70.4]} & 51.1\,\tiny{[45.8, 56.1]} & 92.7\,\tiny{[89.3, 95.6]} \\
\midrule
\multicolumn{4}{l}{\emph{InternVL3.5-8B~\citep{internvl35_2025wang}}} \\
\quad whole prompt & 57.3\,\tiny{[51.1, 62.5]} & 43.2\,\tiny{[38.3, 47.6]} & 86.2\,\tiny{[81.5, 90.4]} \\
\quad pos. per element & 69.2\,\tiny{[63.2, 74.5]} & 55.3\,\tiny{[50.1, 60.1]} & 90.7\,\tiny{[86.4, 94.2]} \\
\quad neg. per element & 64.0\,\tiny{[58.0, 69.4]} & 49.8\,\tiny{[44.8, 54.6]} & 91.0\,\tiny{[87.1, 94.4]} \\
\midrule
\multicolumn{4}{l}{\emph{Qwen2.5-VL-7B~\citep{qwen25vl_2025bai}}} \\
\quad whole prompt & 58.4\,\tiny{[52.1, 64.0]} & 44.6\,\tiny{[39.3, 49.4]} & 87.7\,\tiny{[82.5, 92.4]} \\
\quad pos. per element & 56.3\,\tiny{[49.8, 62.5]} & 43.7\,\tiny{[38.3, 48.9]} & 90.8\,\tiny{[87.6, 93.6]} \\
\quad neg. per element & 57.1\,\tiny{[50.6, 62.7]} & 44.2\,\tiny{[38.8, 49.0]} & 91.3\,\tiny{[88.2, 93.9]} \\
\midrule
\multicolumn{4}{l}{\emph{VQAScore~\citep{vqascore_2024lin}}} \\
\quad whole prompt & 52.3\,\tiny{[45.6, 58.2]} & 38.8\,\tiny{[33.6, 43.7]} & 83.2\,\tiny{[78.0, 88.1]} \\
\quad pos. per element & 55.8\,\tiny{[49.0, 61.6]} & 42.3\,\tiny{[36.9, 47.1]} & 85.8\,\tiny{[80.6, 90.3]} \\
\quad neg. per element & 56.1\,\tiny{[49.5, 62.1]} & 42.8\,\tiny{[37.5, 47.8]} & 86.8\,\tiny{[81.6, 91.4]} \\
\midrule
\multicolumn{4}{l}{\emph{mPLUG~\citep{mplug_2022li}}} \\
\quad whole prompt & 28.9\,\tiny{[21.0, 36.7]} & 21.1\,\tiny{[15.0, 27.1]} & 69.1\,\tiny{[62.8, 74.9]} \\
\quad pos. per element & -5.2\,\tiny{[-13.4, 3.1]} & -4.3\,\tiny{[-10.2, 1.9]} & 39.1\,\tiny{[33.2, 45.3]} \\
\quad neg. per element & -6.9\,\tiny{[-15.2, 1.5]} & -5.5\,\tiny{[-11.5, 0.5]} & 38.4\,\tiny{[32.7, 44.3]} \\
\midrule
\multicolumn{4}{l}{\emph{Similarity and reward}} \\
\quad CLIPScore & 18.0\,\tiny{[9.8, 26.2]} & 13.1\,\tiny{[7.2, 19.1]} & 73.5\,\tiny{[67.6, 78.7]} \\
\quad BLIPScore & 17.0\,\tiny{[8.4, 25.2]} & 12.4\,\tiny{[6.2, 18.4]} & 69.8\,\tiny{[63.5, 75.6]} \\
\quad NegationCLIP & 22.9\,\tiny{[14.6, 30.4]} & 16.6\,\tiny{[10.6, 22.1]} & 74.6\,\tiny{[69.1, 79.8]} \\
\quad ImageReward & 25.7\,\tiny{[17.3, 33.7]} & 18.6\,\tiny{[12.3, 24.8]} & 80.2\,\tiny{[74.5, 85.2]} \\
\quad PickScore & 37.5\,\tiny{[30.0, 44.5]} & 27.7\,\tiny{[22.0, 33.2]} & 82.9\,\tiny{[78.3, 87.0]} \\
\quad HPSv2 & 30.7\,\tiny{[22.7, 37.9]} & 22.3\,\tiny{[16.6, 27.7]} & 80.8\,\tiny{[75.9, 85.2]} \\
\midrule
\bench (ours) & 72.8\,\tiny{[67.9, 77.9]} & \textbf{65.1}\,\tiny{[60.4, 70.0]} & 91.7\,\tiny{[88.5, 95.5]} \\
\midrule
\textcolor{gray}{H. Inter-annotator} & \textcolor{gray}{77.9} & \textcolor{gray}{71.1} & \textcolor{gray}{95.0} \\
\bottomrule
\end{tabular}
\end{table}

\begin{itemize}
\item \textbf{Similarity scores} (CLIPScore~\citep{clipscore_2021hessel}, BLIPScore~\citep{blip_2022li}, NegationCLIP). Cosine similarity between the image embedding and the embedding of the full prompt. NegationCLIP is the negation-aware CLIP of \citet{negationclip_2025park} (ViT-B/32).
\item \textbf{Reward models} (ImageReward~\citep{imagereward_2023xu}, PickScore~\citep{pickscore_2023kirstain}, HPSv2~\citep{hpsv2_2023wu}). The scalar the model assigns to the image and prompt. These models predict which image a person prefers, not whether the image satisfies the prompt.
\item \textbf{VQA judge, whole prompt.} The probability of ``yes'' to one question asking whether the image matches the whole description.
\item \textbf{VQA judge, per element (pos.).} The judge answers one yes/no question per required object and one per statement, all asked in positive form, as in the annotation interface (Figure~\ref{fig:annotator-view}). The expected answer is ``yes'' for positive statements and ``no'' for negated ones. The score is the mean probability of the expected answer over the elements; elements of an absent object count as unsatisfied.
\item \textbf{VQA judge, negated per element (neg.).} The same elements as in pos., with each negated statement asked in negated form and the expected answer flipped. Object questions and positive statements are reused from the pos.\ mode, so the two modes differ only in the wording of the negated statements.
\item \textbf{\bench.} The fraction of elements its detectors mark satisfied, on the same scale as the per-element judges and the human score.
\end{itemize}

The per-element modes give the judges the decomposition that \bench produces. Without it, a judge must either answer about the whole prompt at once, which generally gives lower binary agreement (Table~\ref{tab:sup-binary-all}), or invent its own decomposition. \bench also reports which requirement failed and why, which enables the category and polarity analyses of Section~\ref{sec:experiments}.

\subsection{Measures}
\label{sec:sup-measures}

\textbf{Binary.} Accuracy and Cohen's $\kappa$ against the binary human label. We exclude the similarity and reward scores, because no single threshold applies equally to them.

\textbf{Continuous.} Spearman $\rho$, Kendall $\tau$ and pairwise ranking accuracy against the continuous human score. We report all three because they order the evaluators differently: \bench leads on $\tau$ ($65.1$ vs.\ $63.3$ for Qwen3-VL-32B) but trails on $\rho$ ($72.8$ vs.\ $76.5$) and pairwise accuracy ($91.7$ vs.\ $94.7$). Kendall counts every discordant pair equally, while Spearman penalizes large rank shifts more, so \bench makes more small ranking errors and the judges make fewer but larger ones. The three metrics treat ties and rank displacements differently, so they order the evaluators differently; we report all three.

\textbf{Uncertainty.} All intervals are 95\% percentile intervals from a bootstrap that resamples whole images, so correlated decisions within an image stay together. We compute differences between two evaluators on the same resample, a stronger test than comparing two separate intervals.

\textbf{Similarity and reward scores.} Table~\ref{tab:sup-continuous} includes the six similarity and reward scores. All six rank images far below the VQA judges and \bench: the best, PickScore, reaches $\tau=27.7$, against $38.8$ for the weakest VQA judge other than mPLUG. NegationCLIP, trained on negated captions, improves on CLIPScore only from $13.1$ to $16.6$. Neither similarity nor preference tells whether a negated requirement holds.

\subsection{Cost protocol}
\label{sec:sup-cost}

We measure every method on the same 600 images on one H200 NVL GPU, with one process, no batching, and eight CPU threads. Pinning the threads changes wall-clock time by a factor of $1.2$ and CPU time by a factor of $8$, but changes no verdict. Reported time excludes model loading, which Table~\ref{tab:sup-cost-tab} lists separately. GPU memory is \texttt{torch.cuda.max\_memory\_allocated} for every method, so all values share one basis; the allocator's reserved memory is larger and not comparable across frameworks. CPU memory is the peak resident set size. For \bench we time the full evaluation of each image. For the judges we time each question and report the time per image as the time per question multiplied by the mean number of element questions per image, $4.66$; Table~\ref{tab:sup-cost-tab} gives both values.

\bench has four components: SigLIP 2 SO400M/14 at 378 px for color and material (1,136.0M parameters), SAM~2.1 hiera-small for mask refinement ($38.5$M on the image path, of $46.1$M loaded; the memory-attention modules serve video tracking and are unused), RF-DETR-Seg-XL for detection, counting and masks ($38.1$M), and Depth-Anything-V2-Small for depth relations ($24.8$M). Together they total $1.24$B parameters and $2.0$GiB on disk.

VQAScore and Qwen2.5-VL-7B run through t2v-metrics, which computes a $K\times K$ image-text matrix and keeps only its diagonal. Their time reflects that wrapper as much as the model, so we report it but exclude both from every speed comparison in the paper.

A judge answers one question per element, so its time per image grows linearly with the number of elements. \bench runs its detector once per image. At the benchmark's mean of $4.66$ elements per image, \bench ($0.78$\,s) is faster than the two largest judges, Qwen3-VL-32B ($0.99$\,s) and InternVL3.5-38B ($3.29$\,s), but slower than the two 8B judges with comparable agreement, Qwen3-VL-8B ($0.49$\,s) and InternVL3.5-8B ($0.76$\,s), and uses $2.7$ to $11.6$ times less GPU memory than all of them.

\begin{table}[t]
\centering
\small
\caption{\textbf{Cost of one evaluation.} The same 600 images on one H200 NVL, one process, no batching, eight CPU threads. Model load is reported separately and excluded from s/img. GPU memory is \texttt{torch.cuda.max\_memory\_allocated}. For a judge, s/img is s/Q multiplied by the $4.66$ element questions an image carries on average: every judge is given the benchmark's own decomposition, so they answer identical questions and differ only in s/Q. \bench asks no questions and is timed per image. $^{\dagger}$Scored through t2v-metrics, which computes a $K\times K$ image--text matrix and keeps only its diagonal, so the wall-clock reflects that wrapper rather than the model.}
\label{tab:sup-cost-tab}
\begin{tabular}{lrrrrr}
\toprule
Evaluator & Params & s/Q & s/img & Load (s) & GPU GiB \\
\midrule
InternVL3.5-38B & 38B & 0.706 & 3.29 & 28.6 & 73.0 \\
Qwen3-VL-32B & 32B & 0.212 & 0.99 & 21.8 & 62.8 \\
VQAScore$^{\dagger}$ & 11B & 0.729 & 3.40 & 41.6 & 24.2 \\
Qwen3-VL-8B & 8B & 0.105 & 0.49 & 12.8 & 16.8 \\
InternVL3-8B & 8B & 0.152 & 0.71 & 26.3 & 15.8 \\
InternVL3.5-8B & 8B & 0.164 & 0.76 & 13.8 & 17.1 \\
Qwen2.5-VL-7B$^{\dagger}$ & 7B & 1.803 & 8.39 & 30.3 & 15.9 \\
mPLUG & 0.6B & 0.068 & 0.32 & 22.1 & 6.5 \\
\bench (ours) & 1.24B & -- & 0.78 & 24.5 & 6.3 \\
\bottomrule
\end{tabular}
\end{table}

\subsection{Error structure}
\label{sec:sup-confusion}

Cohen's $\kappa$ merges two kinds of error into one number. Table~\ref{tab:sup-confusion} separates them for the per-element mode, the mode behind the precision and recall of Table~\ref{tab:val-main}.
TP: the annotators and the evaluator both accept the image; TN: both reject it; FP: the annotators reject it but the evaluator passes it; FN: the annotators accept it but the evaluator rejects it. Prec.\ $=\mathrm{TP}/(\mathrm{TP}+\mathrm{FP})$ and Rec.\ $=\mathrm{TP}/(\mathrm{TP}+\mathrm{FN})$.
Called correct (\%) corresponds to the share of the 600 images the evaluator passes, $(\mathrm{TP}+\mathrm{FP})/600$.
Humans accept $34.3\%$ of the images.

\bench accepts $36.8\%$, making $46$ false passes and $31$ false failures, while the strongest judge, Qwen3-VL-32B, accepts $33.5\%$ and makes $32$ and $37$. The two evaluators reject about as many images that humans accept, and \bench passes more images that humans reject. \bench therefore rarely fails an image that humans consider correct, so the generator failures we report are not caused by a strict evaluator.
The same table exposes a degenerate baseline that its aggregate score hides: mPLUG, scored per element, accepts only $8.8\%$ of images and makes $170$ false failures, close to a constant ``no''.

\begin{table}[t]
\centering
\small
\caption{\textbf{Error structure.} Counts against the majority label of three annotators on 600 images, for the per-element scoring mode. FP is an image the annotators reject but the evaluator passes; FN is an image the annotators accept but the evaluator rejects. Judges are ordered by Cohen's $\kappa$; \bench is listed separately.}
\label{tab:sup-confusion}
\begin{tabular}{lrrrrrrr}
\toprule
Evaluator & TP & TN & FP & FN & Prec. & Rec. & Called correct (\%) \\
\midrule
Qwen3-VL-32B, pos. & 169 & 362 & 32 & 37 & \textbf{84.1} & 82.0 & 33.5 \\
Qwen3-VL-8B, pos. & 161 & 358 & 36 & 45 & 81.7 & 78.2 & 32.8 \\
InternVL3.5-38B, pos. & 162 & 356 & 38 & 44 & 81.0 & 78.6 & 33.3 \\
InternVL3.5-8B, pos. & 169 & 346 & 48 & 37 & 77.9 & 82.0 & 36.2 \\
InternVL3-8B, pos. & 148 & 344 & 50 & 58 & 74.7 & 71.8 & 33.0 \\
Qwen2.5-VL-7B, pos. & 117 & 364 & 30 & 89 & 79.6 & 56.8 & 24.5 \\
VQAScore, pos. & 134 & 332 & 62 & 72 & 68.4 & 65.0 & 32.7 \\
mPLUG, pos. & 36 & 377 & 17 & 170 & 67.9 & 17.5 & 8.8 \\
\midrule
\bench (ours) & 175 & 348 & 46 & 31 & 79.2 & \textbf{85.0} & 36.8 \\
\midrule
\textcolor{gray}{Human reference} & \textcolor{gray}{--} & \textcolor{gray}{--} & \textcolor{gray}{--} & \textcolor{gray}{--} & \textcolor{gray}{--} & \textcolor{gray}{--} & \textcolor{gray}{34.3} \\
\bottomrule
\end{tabular}
\end{table}

\subsection{Sensitivity to how a requirement is phrased}
\label{sec:sup-negation}

The pos.\ and neg.\ columns of Table~\ref{tab:val-main} report $\kappa$ on the benchmark's own elements and on the same elements with the negated statements asked in negated form. Every judge loses agreement when using negated prompts. The six instruction-tuned judges, from two model families, lose $7.8$ to $18.8$ points, and scale does not protect them: the two largest judges lose $7.8$ and $16.9$. \bench checks a typed requirement rather than a sentence, so it is unaffected, and under negated phrasing it ranks first.

Table~\ref{tab:sup-negation-probe} shows the mechanism. For each element we ask both the affirmative and the negated form and measure how often the answer flips. A judge that reads the negation should flip every time. Qwen3-VL-32B flips on $77.5\%$ of elements. Qwen2.5-VL-7B flips on only $40.7\%$, and its complementarity error of $0.589$ shows that it mostly repeats its affirmative answer.

\begin{table}[t]
\centering
\small
\setlength{\tabcolsep}{4pt}
\caption{\textbf{Negated-question probe.} Acc.\ whole, pos., neg.: image accuracy against the binary human label when the judge is asked the whole prompt at once, the per-element questions in positive form, or the same questions with the negated statements asked in negated form. Flip: share of negated statements whose yes/no answer changes between the positive and negated wording. Compl.\ err.: mean $|p_{\text{pos}}+p_{\text{neg}}-1|$ over those statements, where $p$ is the probability of ``yes'' under each wording; it is zero when the two answers are complementary. Called corr.\ pos., neg.: share of the 600 images the evaluator passes in each mode; the annotators pass 34.3\%. \textsc{NegT2IBench} has one scoring mode and no question to rephrase.}
\label{tab:sup-negation-probe}
\resizebox{\textwidth}{!}{\begin{tabular}{@{}lrrrrrrr@{}}
\toprule
Evaluator & Acc. whole & Acc. pos. & Acc. neg. & Flip (\%) & Compl. err. & Called corr. pos. (\%) & Called corr. neg. (\%) \\
\midrule
VQAScore & 64.8 & 77.7 & 76.7 & \textbf{80.0} & 0.160 & 32.7 & 35.3 \\
Qwen3-VL-32B & 73.5 & \textbf{88.5} & \textbf{84.7} & 77.5 & 0.225 & 33.5 & 36.3 \\
Qwen3-VL-8B & 74.2 & 86.5 & 77.8 & 62.1 & 0.364 & 32.8 & 44.5 \\
InternVL3.5-8B & 64.7 & 85.8 & 76.5 & 58.9 & 0.361 & 36.2 & 41.2 \\
InternVL3.5-38B & 73.2 & 86.3 & 77.7 & 51.1 & 0.381 & 33.3 & 40.7 \\
InternVL3-8B & 56.8 & 82.0 & 71.5 & 46.7 & 0.431 & 33.0 & 46.8 \\
mPLUG & 46.5 & 68.8 & 67.3 & 46.3 & 0.330 & 8.8 & 6.3 \\
Qwen2.5-VL-7B & 66.2 & 80.2 & 78.3 & 40.7 & 0.589 & 24.5 & 14.7 \\
\bench (ours) & \textbf{87.2} & -- & -- & -- & -- & 36.8 & -- \\
\textcolor{gray}{H. Inter-annotator} & \textcolor{gray}{85.8} & -- & -- & -- & -- & \textcolor{gray}{34.3} & -- \\
\bottomrule
\end{tabular}}
\end{table}

\subsection{Agreement by polarity setting}
\label{sec:sup-by-setting}

Table~\ref{tab:sup-setting-kendall} breaks down Kendall $\tau$ by polarity setting. Each setting holds 70 to 90 images, so a per-cell 95\% interval spans about 32 points and few per-cell differences are resolvable; we read the table only for patterns shared by many evaluators. No strong evaluator collapses in any setting, and \bench stays between $48.1$ and $88.6$. The similarity and reward scores do collapse once the prompt contains a single negation: all six fall from $\tau$ between $20.5$ and $49.1$ at \texttt{1:0} to between $-22.2$ and $20.7$ at \texttt{0:1}.

\begin{table}[t!]
\centering
\small
\setlength{\tabcolsep}{3pt}
\caption{Kendall $\tau$ against the continuous human score, by polarity setting (positive:negated statements). Judges are shown in their three scoring modes. The inter-annotator score (H. inter-annotator) is recomputed within each setting. Settings with no negated statement (1:0, 2:0) have nothing to rephrase, so their negated row is omitted. Each cell holds 70 to 90 images; per-cell 95\% bootstrap intervals have a median width of 32 points. \emph{All} is computed on all 600 images.}
\label{tab:sup-setting-kendall}
\begin{tabular}{@{}lccccccccc@{}}
\toprule
 & \multicolumn{2}{c}{Load 1} & \multicolumn{3}{c}{Load 2} & \multicolumn{2}{c}{Load 3} & \multicolumn{1}{c}{Load 4} & \\
\cmidrule(lr){2-3} \cmidrule(lr){4-6} \cmidrule(lr){7-8} \cmidrule(lr){9-9}
Evaluator & 1:0 & 0:1 & 2:0 & 1:1 & 0:2 & 2:1 & 1:2 & 2:2 & All \\
\midrule
\multicolumn{10}{l}{\emph{Qwen3-VL-32B~\citep{qwen3vl_2025qwen}}} \\
\quad whole prompt & 46.4 & 57.0 & 43.8 & 50.3 & 38.8 & 42.9 & 49.0 & 30.3 & 48.6 \\
\quad pos. per element & \textbf{66.8} & 66.8 & 62.3 & 65.4 & \textbf{67.4} & 60.2 & 51.7 & 57.7 & 63.3 \\
\quad neg. per element & -- & 68.5 & -- & 55.5 & 53.8 & 57.9 & 51.5 & 49.6 & 58.8 \\
\midrule
\multicolumn{10}{l}{\emph{Qwen3-VL-8B~\citep{qwen3vl_2025qwen}}} \\
\quad whole prompt & 54.3 & 55.2 & 39.6 & 32.5 & 40.1 & 38.1 & 47.3 & 39.2 & 46.4 \\
\quad pos. per element & 61.7 & 72.8 & 53.3 & 63.7 & 58.0 & \textbf{64.6} & \textbf{58.8} & 53.9 & 61.0 \\
\quad neg. per element & -- & 57.2 & -- & 47.6 & 49.3 & 62.0 & 55.8 & 48.3 & 54.8 \\
\midrule
\multicolumn{10}{l}{\emph{InternVL3.5-38B~\citep{internvl35_2025wang}}} \\
\quad whole prompt & 52.0 & 57.7 & 39.6 & 38.5 & 44.0 & 49.8 & 43.0 & 32.6 & 49.2 \\
\quad pos. per element & 62.7 & 69.7 & 55.6 & 63.0 & 63.4 & 50.4 & 44.1 & 59.3 & 59.7 \\
\quad neg. per element & -- & 62.6 & -- & 49.4 & 20.2 & 46.6 & 42.0 & 49.8 & 52.8 \\
\midrule
\multicolumn{10}{l}{\emph{InternVL3-8B~\citep{internvl3_zhu2025}}} \\
\quad whole prompt & 47.5 & 55.4 & 42.4 & 43.4 & 22.9 & 47.6 & 36.7 & 38.5 & 46.2 \\
\quad pos. per element & 65.1 & 74.1 & 40.0 & 57.6 & 54.7 & 51.5 & 50.6 & 52.6 & 56.3 \\
\quad neg. per element & -- & 74.4 & -- & 44.4 & 31.5 & 39.9 & 42.1 & 36.6 & 51.1 \\
\midrule
\multicolumn{10}{l}{\emph{InternVL3.5-8B~\citep{internvl35_2025wang}}} \\
\quad whole prompt & 58.3 & 51.8 & 49.0 & 30.9 & 37.1 & 35.1 & 40.7 & 36.1 & 43.2 \\
\quad pos. per element & 57.9 & 63.0 & 63.7 & 55.2 & 57.2 & 45.8 & 36.4 & 45.3 & 55.3 \\
\quad neg. per element & -- & 63.4 & -- & 36.7 & 34.7 & 41.3 & 27.0 & 33.3 & 49.8 \\
\midrule
\multicolumn{10}{l}{\emph{Qwen2.5-VL-7B~\citep{qwen25vl_2025bai}}} \\
\quad whole prompt & 46.9 & 58.2 & 42.4 & 49.0 & 19.8 & 46.3 & 45.9 & 35.5 & 44.6 \\
\quad pos. per element & 51.7 & 58.9 & 45.2 & 55.6 & 40.2 & 34.1 & 38.9 & 43.4 & 43.7 \\
\quad neg. per element & -- & 66.5 & -- & 55.4 & 23.5 & 33.7 & 39.5 & 38.6 & 44.2 \\
\midrule
\multicolumn{10}{l}{\emph{VQAScore~\citep{vqascore_2024lin}}} \\
\quad whole prompt & 36.5 & 50.2 & 26.8 & 10.3 & 23.2 & 48.2 & 48.6 & 24.8 & 38.8 \\
\quad pos. per element & 54.3 & 57.2 & 42.5 & 34.2 & 10.9 & 34.5 & 19.7 & 41.9 & 42.3 \\
\quad neg. per element & -- & 50.0 & -- & 33.8 & 7.8 & 36.5 & 33.3 & 42.6 & 42.8 \\
\midrule
\multicolumn{10}{l}{\emph{mPLUG~\citep{mplug_2022li}}} \\
\quad whole prompt & 37.1 & 18.7 & 22.7 & -0.5 & 1.4 & 11.8 & 15.1 & 4.6 & 21.1 \\
\quad pos. per element & -4.8 & 12.3 & -7.3 & -29.0 & -35.8 & 0.7 & -15.3 & -9.8 & -4.3 \\
\quad neg. per element & -- & 7.3 & -- & -28.1 & -36.0 & -0.2 & -8.3 & -9.6 & -5.5 \\
\midrule
\multicolumn{10}{l}{\emph{Similarity and reward}} \\
\quad CLIPScore & 38.5 & -1.3 & 22.7 & 10.2 & 9.1 & 17.1 & 29.8 & 29.7 & 13.1 \\
\quad BLIPScore & 23.0 & -22.2 & 12.2 & 23.5 & 2.7 & 17.3 & 23.4 & 42.8 & 12.4 \\
\quad NegationCLIP & 29.4 & 2.9 & 30.7 & 22.5 & 0.7 & 14.7 & 17.6 & 36.3 & 16.6 \\
\quad ImageReward & 20.5 & -11.1 & 33.3 & 29.9 & 17.1 & 30.4 & 30.4 & 31.8 & 18.6 \\
\quad PickScore & 49.1 & -1.9 & 32.3 & 41.7 & 6.6 & 23.9 & 23.1 & 35.7 & 27.7 \\
\quad HPSv2 & 38.3 & 20.7 & 27.1 & 33.5 & -0.6 & 23.1 & 34.6 & 22.2 & 22.3 \\
\midrule
\bench (ours) & 62.3 & \textbf{88.6} & \textbf{76.3} & \textbf{77.4} & 66.8 & 48.1 & 50.1 & \textbf{63.2} & \textbf{65.1} \\
\midrule
\textcolor{gray}{H. Inter-annotator} & \textcolor{gray}{79.1} & \textcolor{gray}{88.1} & \textcolor{gray}{73.6} & \textcolor{gray}{66.7} & \textcolor{gray}{76.4} & \textcolor{gray}{65.2} & \textcolor{gray}{69.2} & \textcolor{gray}{73.9} & \textcolor{gray}{71.1} \\
\midrule
\textcolor{gray}{$n$} & \textcolor{gray}{90} & \textcolor{gray}{90} & \textcolor{gray}{70} & \textcolor{gray}{70} & \textcolor{gray}{70} & \textcolor{gray}{70} & \textcolor{gray}{70} & \textcolor{gray}{70} & \textcolor{gray}{600} \\
\bottomrule
\end{tabular}
\end{table}

\subsection{Agreement by category}
\label{sec:sup-by-category}

Table~\ref{tab:sup-category-kendall} breaks down Kendall $\tau$ by category ($100$ images each; per-cell intervals are about $26$ points wide). Judges asked about the whole prompt agree well on color ($65.4$ to $78.4$) but fall on the four relations ($25.3$ to $52.8$); per-element questions recover part of this gap. \bench is more even across categories ($54.4$ to $78.5$), and trails the strongest judges mainly on color and depth, the category most affected by dead zones (Section~\ref{sec:deadzones}). Similarity and reward scores reach useful agreement only on color.

\begin{table}[t!]
\centering
\setlength{\tabcolsep}{3pt}
\caption{Kendall $\tau$ against the continuous human score, by semantic category. Judges are shown in their three scoring modes. The inter-annotator score (H. inter-annotator) is recomputed within each category. Each cell holds 100 images; per-cell 95\% bootstrap intervals have a median width of 26 points. \emph{All} is computed on all 600 images.}
\label{tab:sup-category-kendall}
\small
\begin{tabular}{@{}lccccccc@{}}
\toprule
 & \multicolumn{2}{c}{Attributes} & \multicolumn{4}{c}{Relations} & \\
\cmidrule(lr){2-3} \cmidrule(lr){4-7}
Evaluator & Color & Material & Comparative & Depth & Directional & Proximity & All \\
\midrule
\multicolumn{8}{l}{\emph{Qwen3-VL-32B~\citep{qwen3vl_2025qwen}}} \\
\quad whole prompt & 69.7 & 58.9 & 47.2 & 39.9 & 47.9 & 38.2 & 48.6 \\
\quad pos. per element & 83.2 & 69.7 & 57.3 & \textbf{63.7} & 61.7 & 53.2 & 63.3 \\
\quad neg. per element & 83.5 & 67.8 & 41.9 & 59.7 & 53.2 & 50.1 & 58.8 \\
\midrule
\multicolumn{8}{l}{\emph{Qwen3-VL-8B~\citep{qwen3vl_2025qwen}}} \\
\quad whole prompt & 69.8 & 56.6 & 52.8 & 35.3 & 42.1 & 36.6 & 46.4 \\
\quad pos. per element & \textbf{83.8} & 62.7 & 55.1 & 60.4 & 56.4 & \textbf{61.2} & 61.0 \\
\quad neg. per element & 75.5 & 60.0 & 42.7 & 50.7 & 47.6 & 58.1 & 54.8 \\
\midrule
\multicolumn{8}{l}{\emph{InternVL3.5-38B~\citep{internvl35_2025wang}}} \\
\quad whole prompt & 74.6 & 53.5 & 47.2 & 37.3 & 48.5 & 39.0 & 49.2 \\
\quad pos. per element & 82.0 & 57.3 & 61.7 & 60.5 & 59.0 & 45.1 & 59.7 \\
\quad neg. per element & 83.4 & 59.5 & 40.0 & 44.4 & 49.0 & 50.1 & 52.8 \\
\midrule
\multicolumn{8}{l}{\emph{InternVL3-8B~\citep{internvl3_zhu2025}}} \\
\quad whole prompt & 75.2 & 48.2 & 43.0 & 36.2 & 47.2 & 36.1 & 46.2 \\
\quad pos. per element & 82.3 & 50.6 & 48.3 & 59.3 & 59.7 & 47.0 & 56.3 \\
\quad neg. per element & 82.0 & 52.4 & 47.1 & 42.2 & 36.0 & 51.1 & 51.1 \\
\midrule
\multicolumn{8}{l}{\emph{InternVL3.5-8B~\citep{internvl35_2025wang}}} \\
\quad whole prompt & 71.3 & 45.1 & 46.0 & 35.3 & 39.1 & 36.8 & 43.2 \\
\quad pos. per element & 78.8 & 58.7 & 48.7 & 44.9 & 58.4 & 40.9 & 55.3 \\
\quad neg. per element & 70.6 & 64.3 & 27.8 & 36.3 & 42.3 & 49.5 & 49.8 \\
\midrule
\multicolumn{8}{l}{\emph{Qwen2.5-VL-7B~\citep{qwen25vl_2025bai}}} \\
\quad whole prompt & 78.4 & 60.3 & 37.7 & 33.6 & 42.2 & 39.3 & 44.6 \\
\quad pos. per element & 66.3 & 51.0 & 31.7 & 36.8 & 41.3 & 34.5 & 43.7 \\
\quad neg. per element & 74.1 & 54.5 & 36.3 & 32.2 & 39.7 & 36.3 & 44.2 \\
\midrule
\multicolumn{8}{l}{\emph{VQAScore~\citep{vqascore_2024lin}}} \\
\quad whole prompt & 65.4 & 46.5 & 25.3 & 44.6 & 29.1 & 37.3 & 38.8 \\
\quad pos. per element & 82.4 & 48.8 & 30.0 & 28.3 & 31.9 & 36.0 & 42.3 \\
\quad neg. per element & 82.3 & 44.3 & 30.3 & 36.1 & 32.2 & 42.3 & 42.8 \\
\midrule
\multicolumn{8}{l}{\emph{mPLUG~\citep{mplug_2022li}}} \\
\quad whole prompt & 64.4 & -12.4 & 11.3 & 13.2 & 5.3 & 26.9 & 21.1 \\
\quad pos. per element & 15.9 & 9.2 & -20.3 & -5.4 & -23.1 & -3.4 & -4.3 \\
\quad neg. per element & 15.5 & 3.6 & -20.5 & -4.4 & -20.1 & -6.7 & -5.5 \\
\midrule
\multicolumn{8}{l}{\emph{Similarity and reward}} \\
\quad CLIPScore & 45.9 & -4.0 & 9.0 & 17.2 & 21.1 & 18.8 & 13.1 \\
\quad BLIPScore & 20.0 & -2.4 & 15.6 & 13.9 & 20.2 & 11.7 & 12.4 \\
\quad NegationCLIP & 46.1 & -1.8 & 15.3 & 15.1 & 17.3 & 14.2 & 16.6 \\
\quad ImageReward & 38.1 & -8.6 & 23.2 & 28.8 & 23.0 & 15.0 & 18.6 \\
\quad PickScore & 56.6 & 11.2 & 34.8 & 23.3 & 25.5 & 10.6 & 27.7 \\
\quad HPSv2 & 44.2 & 25.3 & 27.7 & 13.1 & 9.1 & 18.3 & 22.3 \\
\midrule
\bench (ours) & 75.2 & \textbf{78.5} & \textbf{72.3} & 54.4 & \textbf{65.3} & 60.0 & \textbf{65.1} \\
\midrule
\textcolor{gray}{H. Inter-annotator} & \textcolor{gray}{84.2} & \textcolor{gray}{81.4} & \textcolor{gray}{78.8} & \textcolor{gray}{66.5} & \textcolor{gray}{65.8} & \textcolor{gray}{64.3} & \textcolor{gray}{71.1} \\
\bottomrule
\end{tabular}
\end{table}

\subsection{Per-element agreement}
\label{sec:sup-elements}

Table~\ref{tab:sup-elements-polarity} scores each object and statement on its own instead of one verdict per image, so a negated statement is judged even when its image fails on another element. Annotators also agree less on negated statements, so each block has its own inter-annotator score (73.3 positive, 59.7 negated).
On positive statements the strongest judges agree more closely than \bench (Qwen3-VL-32B 76.3 vs.\ 64.3). On negated statements the gap narrows: \bench reaches $\kappa=56.4$, within the interval of the best judge (60.7) and close to the inter-annotator score (59.7), and it loses less agreement from positive to negated statements than the three strongest judges (7.9 points, against 14.3 to 18.5).
Asked in negated form, every judge falls further, to between $2.3$ and $43.5$.

\begin{table}[t!]
\centering
\small
\caption{\textbf{Element-level agreement, by polarity.} One decision per object and per statement instead of one verdict per image. Negated statements are harder for the annotators too, so each block has its own inter-annotator score. The last three columns give the share of satisfied and of unsatisfied statements each method gets right, and their mean. Intervals on $\kappa$ are 95\% bootstrap intervals resampling images.}
\label{tab:sup-elements-polarity}
\begin{tabular}{lrrrrrr}
\toprule
Evaluator & $n$ & Acc. & $\kappa$ & Satisfied & Unsatisfied & Balanced \\
\midrule
\multicolumn{7}{l}{\emph{Affirmative statements}} \\
\bench (ours) & 2249 & 90.5 & 64.3\,\tiny{[59.6, 68.9]} & 95.6 & 65.5 & 80.5 \\
VQAScore, pos. & 2242 & 88.8 & 53.4\,\tiny{[48.1, 58.4]} & 96.9 & 48.9 & 72.9 \\
mPLUG, pos. & 2242 & 42.5 & -1.0\,\tiny{[-3.9, 2.0]} & 39.2 & 58.8 & 49.0 \\
Qwen2.5-VL-7B, pos. & 2242 & 79.0 & 48.3\,\tiny{[44.4, 52.1]} & 75.9 & 94.4 & 85.1 \\
InternVL3-8B, pos. & 2242 & 90.9 & 63.8\,\tiny{[59.2, 68.6]} & 97.2 & 60.1 & 78.6 \\
Qwen3-VL-8B, pos. & 2242 & 92.6 & 72.9\,\tiny{[69.0, 76.9]} & 95.9 & 75.9 & 85.9 \\
Qwen3-VL-32B, pos. & 2242 & 93.3 & \textbf{76.3}\,\tiny{[72.8, 79.8]} & 95.7 & 81.4 & 88.6 \\
InternVL3.5-8B, pos. & 2242 & 91.2 & 64.9\,\tiny{[60.1, 69.5]} & 97.4 & 60.7 & 79.1 \\
InternVL3.5-38B, pos. & 2242 & 91.7 & 68.1\,\tiny{[63.8, 72.3]} & 96.6 & 67.1 & 81.9 \\
\textcolor{gray}{H. inter-annotator} & \textcolor{gray}{2249} & \textcolor{gray}{91.6} & \textcolor{gray}{73.3} & \textcolor{gray}{--} & \textcolor{gray}{--} & \textcolor{gray}{--} \\
\midrule
\multicolumn{7}{l}{\emph{Negated statements}} \\
\bench (ours) & 523 & 78.2 & 56.4\,\tiny{[48.8, 63.5]} & 76.4 & 80.2 & 78.3 \\
VQAScore, pos. & 519 & 61.3 & 23.7\,\tiny{[15.6, 31.1]} & 41.3 & 82.8 & 62.0 \\
VQAScore, neg. & 519 & 58.8 & 18.3\,\tiny{[9.9, 26.6]} & 46.8 & 71.6 & 59.2 \\
mPLUG, pos. & 519 & 59.3 & 20.0\,\tiny{[11.9, 27.4]} & 37.5 & 82.8 & 60.2 \\
mPLUG, neg. & 519 & 50.7 & 2.3\,\tiny{[-6.2, 10.5]} & 37.2 & 65.2 & 51.2 \\
Qwen2.5-VL-7B, pos. & 519 & 72.4 & 44.0\,\tiny{[37.2, 51.0]} & 92.2 & 51.2 & 71.7 \\
Qwen2.5-VL-7B, neg. & 519 & 59.7 & 21.4\,\tiny{[15.4, 27.9]} & 26.4 & 95.6 & 61.0 \\
InternVL3-8B, pos. & 519 & 73.0 & 46.6\,\tiny{[39.1, 54.0]} & 57.6 & 89.6 & 73.6 \\
InternVL3-8B, neg. & 519 & 54.7 & 8.5\,\tiny{[0.7, 17.2]} & 67.7 & 40.8 & 54.2 \\
Qwen3-VL-8B, pos. & 520 & 77.1 & 54.4\,\tiny{[46.8, 62.4]} & 70.7 & 84.0 & 77.4 \\
Qwen3-VL-8B, neg. & 519 & 64.9 & 28.4\,\tiny{[21.3, 36.9]} & 91.4 & 36.4 & 63.9 \\
Qwen3-VL-32B, pos. & 520 & 80.4 & \textbf{60.7}\,\tiny{[53.8, 67.4]} & 80.7 & 80.0 & 80.4 \\
Qwen3-VL-32B, neg. & 519 & 72.1 & 43.5\,\tiny{[36.3, 51.1]} & 85.5 & 57.6 & 71.6 \\
InternVL3.5-8B, pos. & 520 & 75.0 & 50.4\,\tiny{[43.1, 57.9]} & 64.1 & 86.8 & 75.4 \\
InternVL3.5-8B, neg. & 519 & 58.4 & 16.7\,\tiny{[7.9, 26.1]} & 59.1 & 57.6 & 58.4 \\
InternVL3.5-38B, pos. & 520 & 76.7 & 53.8\,\tiny{[45.9, 61.7]} & 66.3 & 88.0 & 77.1 \\
InternVL3.5-38B, neg. & 519 & 59.9 & 18.9\,\tiny{[10.1, 28.2]} & 75.5 & 43.2 & 59.3 \\
\textcolor{gray}{H. inter-annotator} & \textcolor{gray}{523} & \textcolor{gray}{80.0} & \textcolor{gray}{59.7} & \textcolor{gray}{--} & \textcolor{gray}{--} & \textcolor{gray}{--} \\
\bottomrule
\end{tabular}
\end{table}

\section{Evaluation rules and metrics}
\subsection{Evaluation rules}
\label{app:evaluation-details}
\begin{table}[t]
\small{
\centering
\caption{\textbf{\bench evaluation rules.} Boolean tests on masks by category: positive statements pass when condition holds, negated ones when it does not. Categories span diverse negation forms.}
\label{tab:decision-rules}
{\footnotesize
\setlength{\tabcolsep}{6pt}
\begin{tabular}{@{}p{0.175\linewidth} p{0.18\linewidth} p{0.59\linewidth}@{}}
\toprule
Category (\#values) & What we check & Condition for the attribute or relation to hold \\
\midrule

Color (10) / \newline Material (4) &
The object carries the requested color / material &
$s_v \geq s_{\max} - \beta, (\beta=0.5 \text{ for color, } 0 \text{ for material}) $
\newline
$s_v$ is SigLIP 2 score for color / material $v$ on object crop and
$s_{\max}$ the highest; for color, objects count as having every value within
$0.5$ of $s_{\max}$, so when two values score closely, a statement
denying either fails; material uses the top value only \\
\addlinespace

Directional (4) &
The subject lies on the requested side of the target &
left / right: $d_x < -0.45$, $d_x > 0.45$
\newline
above / below: $d_y < -0.45$, $d_y > 0.45$
\newline
$(d_x,d_y)$ is offset between mask centers, shrunk by
$0.075 \times$ summed mask extents and normalized by original
center distance \\
\addlinespace

Proximity (2) &
Two objects are near each other or far apart &
near: $g < 0.75R$; far from: $g \geq 0.75R$
\newline
$g$ is minimum mask-to-mask gap and $R$ is mean equivalent radius.
Every measurable pair is either near or far, so proximity has no
designed ambiguous middle \\
\addlinespace

Comparative (4) &
One object is bigger or taller than another &
bigger / smaller: $a_A > 1.05a_B$;
$a_B > 1.05a_A$
\newline
taller / shorter: $h_A > 1.05h_B$;
$h_B > 1.05h_A$
\newline
$a$ is mask area and $h$ mask height; pairs within the factor are called
neither way \\
\addlinespace

Depth (2) &
The subject is in front of the target, or behind it &
Let $\bar{\delta}_A$ and $\bar{\delta}_B$ be the mean disparities over the
two objects' facing surfaces, defined within the mask-to-mask gap plus
$16$ pixels. A depth relation is emitted only if
\[
\frac{|\bar{\delta}_A-\bar{\delta}_B|}
     {\delta_{\max}-\delta_{\min}}
\geq
0.05+0.10\frac{g}{R}.
\]
If the condition holds, the object with larger disparity is in front; otherwise
no depth predicate is emitted \\
\bottomrule
\end{tabular}}}
\end{table}

All reported benchmark results use RF-DETR-XL with the current evaluator
configuration. The formal rules below are deterministic. Unless otherwise
noted, only relation measurement operates on a SAM-2-refined
instance mask.

\textbf{Detection and instance counting.}
RF-DETR-XL (\texttt{RFDETRSegXLarge}) is called with an internal prediction
floor of $0.005$. A requested class is considered present when at least one
detection reaches
\[
\tau_{\mathrm{presence}} = 0.03.
\]
Instance counting uses the stricter threshold.
\[
\tau_{\mathrm{count}} = 0.5.
\]
If no detection reaches $\tau_{\mathrm{count}}$, the most confident detection
for that requested class is retained so that a faint but detected object can
still be judged. If more counted instances of a requested class are present
than requested, the prompt is marked as an over-count. \texttt{skis} is exempt
from this rule because one semantic referent is often returned as multiple
detections. At most 16 detections per class are retained, and overlap
suppression is disabled ($\mathrm{IoU}=1.0$).

\textbf{SAM 2 mask refinement.}
We refine detections with
\texttt{facebook/sam2.1-hiera-small}. Detector boxes are passed as SAM box
prompts; the candidate mask is selected by overlap with the detector's own mask
rather than by SAM confidence. The refined mask is post-processed with
morphological closing of radius 5 pixels. Connected components smaller than
$5\%$ of the largest component are removed, while larger disconnected pieces
are retained to preserve partially occluded objects.
SAM-refined masks are used for only the four relation categories. SAM
refinement is restricted to detections corresponding to objects named
in the prompt. Each relation prompt names at most five objects.

\textbf{Attribute decision rule.}
\label{app:attribute-rule}
Color and material are evaluated with
\texttt{ViT-SO400M-14-SigLIP2-378} SigLIP 2 features
(\texttt{pretrained=webli}). Pixels outside the refined instance mask are
filled with gray (\texttt{\#999}) before the instance crop is classified.

Let $V$ be the attribute vocabulary and $s_v$ the SigLIP 2 logit for value
$v\in V$. The evaluator forms the multi-label confusion set.
\[
C(I,o)
=
\left\{
v\in V:
s_v \ge \max_{u\in V}s_u-\beta
\right\},
\qquad
\beta = 0.5\text{ for color and }\beta = 0\text{ for material (plain top-1).}
\]
For a requested value $v^\star$, an asserted attribute passes if and only if
$v^\star\in C(I,o)$,
while a negated attribute passes if and only if
$v^\star\notin C(I,o)$.
Thus, a near-tied denied value is not treated as confidently absent.

The color vocabulary is
\[
\{\text{red, orange, yellow, green, blue, purple, pink, brown, black, white}\},
\]
and the material vocabulary is
\[
\{\text{wooden, metal, plastic, glass}\}.
\]

\textbf{Directional relation rule.}
\label{app:directional-rule}
We adapt GenEval's position rule~\citep{geneval_2023ghosh}, using refined-mask centroids and extents in place of bounding-box centers and sizes, with the thresholds reported below.

Let $M_A,M_B$ be the refined masks, $c_A,c_B$ their centroids, and
$(w_A,h_A),(w_B,h_B)$ their mask extents. With centroid offset
$o=c_A-c_B$, define the extent-discounted offset
\[
o_e
=
\max\!\left(
|o|-\tau_{\mathrm{dir}}
(w_A+w_B,\;h_A+h_B),
0
\right)\odot \operatorname{sign}(o),
\]
with
$\tau_{\mathrm{dir}}=0.075$.
The evaluator then normalizes the discounted offset using the original
centroid-displacement magnitude,
$(d_x,d_y)=\frac{o_e}{\|o\|_2}.$
A corresponding signed directional predicate fires when the magnitude of its
normalized component exceeds
$\alpha_{\mathrm{dir}}=0.45$.
If neither axis clears the threshold, no directional predicate is emitted for
the pair.

\textbf{Proximity relation rule.}
\label{app:proximity-rule}

The minimum edge-to-edge distance between the two refined masks is
\[
g
=
\min_{p\in M_A,\;q\in M_B}
\|p-q\|_2.
\]
Define the mean equivalent-circle radius
\[
R
=
\frac{1}{2}
\left(
\sqrt{\frac{|M_A|}{\pi}}
+
\sqrt{\frac{|M_B|}{\pi}}
\right).
\]
The decision boundary is
\[
\text{near}
\Longleftrightarrow
g < \gamma_{\mathrm{prox}}R,
\qquad
\text{far from}
\Longleftrightarrow
g \ge \gamma_{\mathrm{prox}}R,
\]
with
$\gamma_{\mathrm{prox}}=0.75$.
Near and far are therefore exact complements, so proximity has no geometric
dead zone.

\textbf{Comparative relation rule.}
\label{app:comparative-rule}
Let $a_X=|M_X|$ denote refined-mask area and let $h_X$ denote its vertical
extent. With
$\rho_{\mathrm{size}}=1.05$,
the area predicates are
\[
A\text{ bigger than }B
\Longleftrightarrow
a_A>\rho_{\mathrm{size}}a_B,
\]
\[
A\text{ smaller than }B
\Longleftrightarrow
a_B>\rho_{\mathrm{size}}a_A,
\]
and the height predicates are
\[
A\text{ taller than }B
\Longleftrightarrow
h_A>\rho_{\mathrm{size}}h_B,
\]
\[
A\text{ shorter than }B
\Longleftrightarrow
h_B>\rho_{\mathrm{size}}h_A.
\]
If neither direction exceeds the ratio margin, the evaluator emits no
comparative predicate. These relations measure visible in-image size rather
than real-world physical size.

\textbf{Depth relation rule.}
\label{app:depth-rule}
Depth is estimated with
\texttt{depth-anything/Depth-Anything-V2-Small-hf}, whose disparity output
$\delta$ increases for surfaces closer to the camera.

Let $M_A,M_B$ be the refined masks. Define the mask gap
\[
g
=
\min_{p\in M_A,\;q\in M_B}
\|p-q\|_2
\]
and the pair's mean equivalent-circle radius
\[
R
=
\frac{1}{2}
\left(
\sqrt{\frac{|M_A|}{\pi}}
+
\sqrt{\frac{|M_B|}{\pi}}
\right).
\]
For a fixed facing-window width $w=16$ px, the facing surfaces are
\[
F_A
=
\left\{
p\in M_A:
d(p,M_B)\le g+w
\right\},
\]
\[
F_B
=
\left\{
q\in M_B:
d(q,M_A)\le g+w
\right\},
\]
where $d(\cdot,M)$ is Euclidean distance to the nearest pixel of mask $M$.
Let $\bar{\delta}_A$ and $\bar{\delta}_B$ be the mean disparity over
$F_A$ and $F_B$, and let $\Delta_{\mathrm{img}}$ be the disparity range of the
image. The evaluator commits to a depth predicate only if
\[
\frac{
|\bar{\delta}_A-\bar{\delta}_B|
}{
\Delta_{\mathrm{img}}
}
\ge
\beta_0+\beta_1\frac{g}{R},
\]
where
$\beta_0=0.05$,
$\beta_1=0.10$.
If the condition holds, $\bar{\delta}_A>\bar{\delta}_B$ yields
\emph{$A$ in front of $B$}, while
$\bar{\delta}_A<\bar{\delta}_B$ yields
\emph{$A$ behind $B$}. Otherwise no depth predicate is emitted.

This coupling makes the required depth evidence increase with spatial
separation: touching objects require only the base margin, while distant pairs
must differ more strongly in disparity before the evaluator commits.

\textbf{Positive, negated, and dead-zone semantics.}
\label{app:dead-zones}
For each relation category the evaluator returns the set of predicates that
fire for a given object pair. An asserted statement passes if and only if its requested
predicate belongs to this set; a negated statement passes if and only if its denied
predicate does not.

Directional, comparative, and depth contain a deliberate dead zone in which no
predicate fires because the measured geometry is not sufficiently decisive.
A positive statement fails in such a region, whereas a negated statement
passes because the denied predicate is absent. Proximity has no dead zone by
construction because its two predicates are exact complements. Attribute
confusion sets are also never empty because they always include the top-scoring
value.

\subsection{Evaluator configuration}
\label{app:experiment-evaluator}

Table~\ref{tab:current-evaluator-config} summarizes the evaluator used
for all generator benchmark scores. RF-DETR-XL provides detections and
instance masks; SAM~2 refines masks for relation measurements only.
SigLIP2 scores detector-masked attribute crops without SAM, while
Depth-Anything V2 provides depth estimates. No benchmark verdict uses
a generative language-model judge. The decision rules are detailed in
Appendix~\ref{app:evaluation-details}.

The ablation varies the detector, SAM refinement, and relation-rule
configuration. Table~\ref{tab:previous-evaluator-config} lists the
previous relation constants; this axis leaves the SigLIP2 attribute
configuration unchanged. Detector-specific thresholds and SAM usage
are controlled by their respective ablation axes.

\begin{table*}[t]
\caption{\textbf{Current evaluator configuration used for generator
benchmark scoring.} Attribute buffers are measured in
$100\times$ cosine-similarity units.}
\label{tab:current-evaluator-config}
\centering
\small
\begin{tabular}{lll}
\toprule
Stage & Setting & Current value \\
\midrule
Detection & Backbone & RF-DETR-XL (\texttt{RFDETRSegXLarge}) \\
Detection & Presence threshold & 0.03 \\
Detection & Counting threshold & 0.5 \\
Detection & Per-class cap & 16 \\
Detection & Overlap suppression & disabled ($\mathrm{IoU}=1.0$) \\
Masks & Refiner & SAM 2.1 Hiera Small \\
Masks & Closing radius & 5 px \\
Masks & Minimum component & $0.05\times$ largest component \\
Masks & Refined-mask scope & prompt instances in all four relation categories \\
Masks & Box-prompt chunk size & 50 \\
Attributes & Backbone / weights & ViT-SO400M-14-SigLIP2-378 / webli \\
Attributes & Crop mask & detector mask; no SAM \\
Attributes & Confusion buffer $\beta$ & 0.5 (color), 0 (material) \\
Attributes & Additional color answer & gray (not requested in prompts) \\
Directional & Extent margin $\tau_{\mathrm{dir}}$ & 0.075 \\
Directional & Axis threshold $\alpha_{\mathrm{dir}}$ & 0.45 \\
Proximity & Gap factor $\gamma_{\mathrm{prox}}$ & 0.75 \\
Comparative & Size ratio $\rho_{\mathrm{size}}$ & 1.05 \\
Depth & Rule & facing surfaces with coupled confidence \\
Depth & Facing window $w$ & 16 px \\
Depth & Base margin $\beta_0$ & 0.05 \\
Depth & Separation slope $\beta_1$ & 0.10 \\
\bottomrule
\end{tabular}
\end{table*}

\begin{table*}[t]
\caption{\textbf{Previous relation-rule configuration used in the
evaluator ablation.} These settings replace the corresponding
current relation rules; attribute scoring is unchanged.}
\label{tab:previous-evaluator-config}
\centering
\small
\begin{tabular}{lll}
\toprule
Stage & Setting & Previous value \\
\midrule
Directional & Extent margin $\tau_{\mathrm{dir}}$ & 0.10 \\
Directional & Axis threshold $\alpha_{\mathrm{dir}}$ & 0.50 \\
Proximity & Gap factor $\gamma_{\mathrm{prox}}$ & 0.50 \\
Comparative & Size ratio $\rho_{\mathrm{size}}$ & 1.15 \\
Depth & Rule & contact-band rule \\
Depth & Contact-band width & 12 px \\
Depth & Minimum rim/contact pixels & 20 \\
\bottomrule
\end{tabular}
\end{table*}

\subsection{Metrics and failure definitions}
\label{app:metric-definitions}
\textbf{Metrics.} Let $\mathcal{I}$ be the set of generated images and
$\operatorname{correct}(I)\in\{0,1\}$ denote complete-prompt correctness.

\textbf{Image accuracy} is
\[
\mathrm{Acc}_{\mathrm{img}}
=
\frac{1}{|\mathcal{I}|}
\sum_{I\in\mathcal{I}}
\operatorname{correct}(I).
\]

Let $\mathcal{P}$ denote the prompt set and $\mathcal{I}(p)$ the four images
generated for prompt $p$. 

\textbf{Prompt accuracy} is
\[
\mathrm{Acc}_{\mathrm{prompt}}
=
\frac{1}{|\mathcal{P}|}
\sum_{p\in\mathcal{P}}
\mathbf{1}
\left[
\exists I\in\mathcal{I}(p):
\operatorname{correct}(I)=1
\right].
\]

\textbf{Failure decomposition.}
For diagnosis, the evaluator also records why an incorrect sample failed.
Statement-level failures are assigned to \emph{undetected object},
\emph{over-count}, \emph{positive violated}, \emph{negated violated}, or
\emph{dead-zone failure}; row-level failures distinguish object/count errors,
positive violations, negated violations, or both polarities. These diagnostics
do not define a separate benchmark score.

\section{Evaluator ablations}
\label{app:evaluator-ablations}

\textbf{Compared configurations.}
We ablate three binary design choices: the object detector
(Mask R-CNN~\citep{maskrcnn_2017he} vs.\ RF-DETR-XL~\citep{rfdetr_2026robinson}), SAM refinement (on vs.\ off), and the
relation-rule constants (previous vs.\ current). This yields:
C1 = Mask R-CNN + SAM + previous constants;
C2 = Mask R-CNN + SAM + current constants;
C3 = Mask R-CNN + no SAM + previous constants;
C4 = Mask R-CNN + no SAM + current constants;
C5 = RF-DETR-XL + SAM + previous constants;
C6 = RF-DETR-XL + SAM + current constants;
C7 = RF-DETR-XL + no SAM + previous constants; and
C8 = RF-DETR-XL + no SAM + current constants.
Table~\ref{tab:evaluator-ablation-full} reports agreement with human annotations on
the 140 examples with a clean human-majority verdict. We use C6 throughout
\bench.

\textbf{Threshold selection.}
All configurations are scored with the evaluator from
\S\ref{sec:evaluation-protocol}: RF-DETR-XL at presence/counting thresholds
$0.03/0.5$, SAM~2-refined masks for all relation evaluations,
SigLIP 2 for color/material, and Depth-Anything V2 for depth. The geometric cut
points and attribute confusion buffer were fitted against 140 human-annotated
images and then fixed. The evaluator is deterministic and contains no
language-model judge.

\begin{table*}[t]
\caption{\textbf{Full evaluator ablation against human annotations.}}
\label{tab:evaluator-ablation-full}
\centering
\small
\begin{tabular}{cllccccc}
\toprule
Config. & Detector & SAM & Constants & $n$ &
Agreement (\%) & $\kappa$ \\
\midrule
C6 & RF-DETR-XL & \checkmark & Current  & 140 & 88.5 & 0.732 \\
C8 & RF-DETR-XL & --         & Current  & 140 & 87.1 & 0.698 \\
C2 & Mask R-CNN & \checkmark & Current  & 140 & 85.6 & 0.654 \\
C4 & Mask R-CNN & --         & Current  & 140 & 84.9 & 0.640 \\
C7 & RF-DETR-XL & --         & Previous & 140 & 82.0 & 0.578 \\
C1 & Mask R-CNN & \checkmark & Previous & 140 & 81.3 & 0.557 \\
C3 & Mask R-CNN & --         & Previous & 140 & 81.3 & 0.557 \\
C5 & RF-DETR-XL & \checkmark & Previous & 140 & 81.3 & 0.564 \\
\bottomrule
\end{tabular}
\end{table*}

The ablation was sampled from 140 of 600 annotated images distributed across model, setting, and category; they have a clear human-majority verdict and enter the headline table. The point estimates should therefore be read as a compact diagnostic of evaluator design rather than as a precise ranking of configurations.
SAM refinement is retained in C6 to tighten object masks for geometric
relation measurements, including area, centroid, spacing, and depth
computations.

\section{Generator experiments}
\subsection{Experimental setup}
\label{app:experiments}

\textbf{Models and sample counts.}
We evaluate eleven model configurations on the full \bench corpus:
BLIP3-o NEXT SFT 3B~\citep{blip3onext_2025chen},
FLUX.2-klein-9B and FLUX.2-dev (4-bit)~\citep{flux2_2025bfl},
HiDream-I1 (4-bit)~\citep{hidream_2025cai},
HunyuanImage-2.1 (distilled)~\citep{hunyuanimage_2025tencent},
Janus-Pro-7B~\citep{januspro_2025chen},
Qwen-Image~\citep{qwenimage_2025qwenteam},
SD~3.5 Large~\citep{sd3_2024esser},
Z-Image-Turbo~\citep{zimage_2025team}, and
SD~1.5~\citep{ldm_2022rombach} with and without the negation-aware text
encoder of \citet{negationclip_2025park}.
Each configuration generates four samples for the same 4,800 prompts, yielding
19,200 evaluated images per configuration and 211,200 images in total.

\textbf{Generation settings.}
\label{app:generation-settings}
All model configurations use the same \bench prompt files and four samples per
prompt with seeds $42,\ldots,45$. We intentionally use each pipeline's native
generation configuration rather than forcing an equal inference-compute
budget. Consequently, absolute between-model comparisons also reflect
differences in inference schedule and output resolution; within-model
comparisons across benchmark settings do not.

\begin{table}[t]
\caption{\textbf{Generation settings used in the benchmark sweep.}
Resolutions are pipeline defaults. Janus-Pro-7B is autoregressive, so a
diffusion step count is not applicable.}
\label{tab:generation-settings}
\centering
\begin{tabular}{lcc}
\toprule
Model & Steps & Resolution \\
\midrule
FLUX.2-klein-9B              & 4  & $1024^2$ \\
HunyuanImage-2.1 (distilled) & 8  & $1024^2$ \\
Z-Image-Turbo                & 9  & $1024^2$ \\
FLUX.2-dev (4-bit)           & 28 & $1024^2$ \\
SD~3.5 Large                 & 28 & $1024^2$ \\
BLIP3-o NEXT SFT 3B          & 30 & $1024^2$ \\
HiDream-I1 (4-bit)           & 50 & $1024^2$ \\
Qwen-Image                   & 50 & $1024^2$ \\
SD~1.5                       & 50 & $512^2$ \\
SD~1.5 + NegationCLIP        & 50 & $512^2$ \\
Janus-Pro-7B                 & -- & $384^2$ \\
\bottomrule
\end{tabular}
\end{table}

No model-specific prompt engineering is applied, and the CFG negative-prompt
field is unused. The SD~1.5 and SD~1.5+NegationCLIP entries share the same base
checkpoint and 50-step schedule, but their registered guidance scales are 9.0
and 7.5 respectively; they should therefore not be interpreted as a perfectly
controlled encoder intervention.

\subsection{Additional results}
\label{app:additional-results}
\textbf{Failure analysis.}
\label{app:category-failures}
The main paper reports failure composition by benchmark setting and semantic
category. Here we provide the corresponding statement-count calculation and
additional image- and model-level diagnostics.
Across the full benchmark there are 475{,}200 statement evaluations. For one
prompt from each of the eight settings, the total statement load is
\[
1+1+2+2+2+3+3+4=18.
\]
With six categories, 100 prompts per category--setting cell, four generated
samples per prompt, and eleven model configurations, this gives
\[
6 \times 100 \times 4 \times 11 \times 18 = 475{,}200
\]
statement evaluations. Of these, 280,067 are failed statements; the
statement-level percentages in Section~\ref{sec:failures} are computed over
these failed statements, or over the corresponding failed statements within
each setting or category.

At the image level, 158,421 of 211,200 generations are incorrect. Among
these, 40.1\% contain only negated-statement violations, 25.9\% only
positive-statement violations, and 13.9\% violations of both polarities.
Over-counting accounts for a further 10.5\% and undetected objects for 9.6\%.
Images that violate both positive and negated statements are retained as a
separate category rather than assigned to either polarity.

\textbf{Dead-zone analysis.}
\label{app:deadzones}
Directional, comparative, and depth relations contain a deliberate
\emph{dead zone}: when the measured geometry is insufficient to support any
predicate in the category, the evaluator returns an empty predicate set rather
than forcing a decision. This can be the correct outcome; for example, two
objects at approximately the same depth may be neither clearly
\emph{in front of} nor \emph{behind} one another.

Let $r$ denote the relation predicate a statement names (e.g., \emph{left of}) and $R(A,B)$ the set of predicates the evaluator returns for the object pair $(A,B)$. A positive statement passes if and only if $r \in R(A,B)$, whereas a negated statement passes if and only if $r \notin R(A,B)$.
Consequently, when $R(A,B)=\emptyset$, a positive statement fails while a negated statement passes.
We call the latter a \emph{dead-zone pass}. Such a
pass is valid under the benchmark rule, but differs from a pass in which an
alternative predicate is positively established.

\begin{table}[t]
\caption{\textbf{Dead-zone exposure among successful negated relations.}
A dead-zone pass occurs when a negated statement succeeds while the relation
evaluator returns no predicate for that object pair.}
\label{tab:deadzone-exposure}
\centering
\begin{tabular}{@{}lrrrr@{}}
\toprule
Category & Evaluable & Passed & Dead-zone passes & Share of passes \\
\midrule
Depth       & 30{,}335 & 17{,}947 & 9{,}963 & \textbf{55.5\%} \\
Directional & 29{,}777 & 17{,}117 & 1{,}567 & 9.2\% \\
Comparative & 29{,}883 & 12{,}962 & 374 & 2.9\% \\
Proximity   & 30{,}620 & 14{,}764 & 3         & 0.0\% \\
\midrule
Total & 120{,}615 & 62{,}790 & 11{,}907 & 19.0\% \\
\bottomrule
\end{tabular}
\end{table}

Across the full experiment, 158,400 negated relation statements are evaluated
across the 211,200 generated images. Of these, shown in Table~\ref{tab:deadzone-exposure}, 37,785 are unevaluable because
an endpoint is missing or over-counted. Among the remaining evaluable
statements, 62,790 pass, and 11,907 of those passes (19.0\%) arise from the
dead zone. Equivalently, dead-zone passes account for 7.5\% of all negated
relation statements.

The effect is concentrated in depth: 55.5\% of successful negated depth
statements occur when no depth predicate is resolved. This does not imply that
those images are incorrectly scored; an object pair may genuinely lack a clear
front--behind ordering. It does mean, however, that such cases provide weaker
evidence of negation following than cases in which the evaluator positively
establishes the alternative relation.

Proximity provides a useful contrast. Its \emph{near}/\emph{far} predicates
are exact complements under a single decision boundary, so it has no designed
dead zone. The three observed empty-set cases are degenerate geometry rather
than a consequence of the decision rule.

\textbf{Negation-aware text encoder.}
Replacing SD~1.5's text encoder with NegationCLIP changes overall image accuracy
only from 14.5\% to 14.9\%. Both SD~1.5 configurations are unusual in scoring
higher on \texttt{pos0\_neg1} than on \texttt{pos1\_neg0}, but the pair remains
well below the other configurations overall. This comparison is not fully
controlled: the checkpoint and 50-step schedule are shared, but the registered
guidance scales differ (9.0 vs.\ 7.5). We therefore treat the pair as suggestive
rather than as an isolated causal test of the text encoder.

\subsection{Additional qualitative examples.}
\label{app:additional-qualitative-examples}
\begin{figure*}[t]
 \centering
 \includegraphics[width=\textwidth,height=\textheight,keepaspectratio]{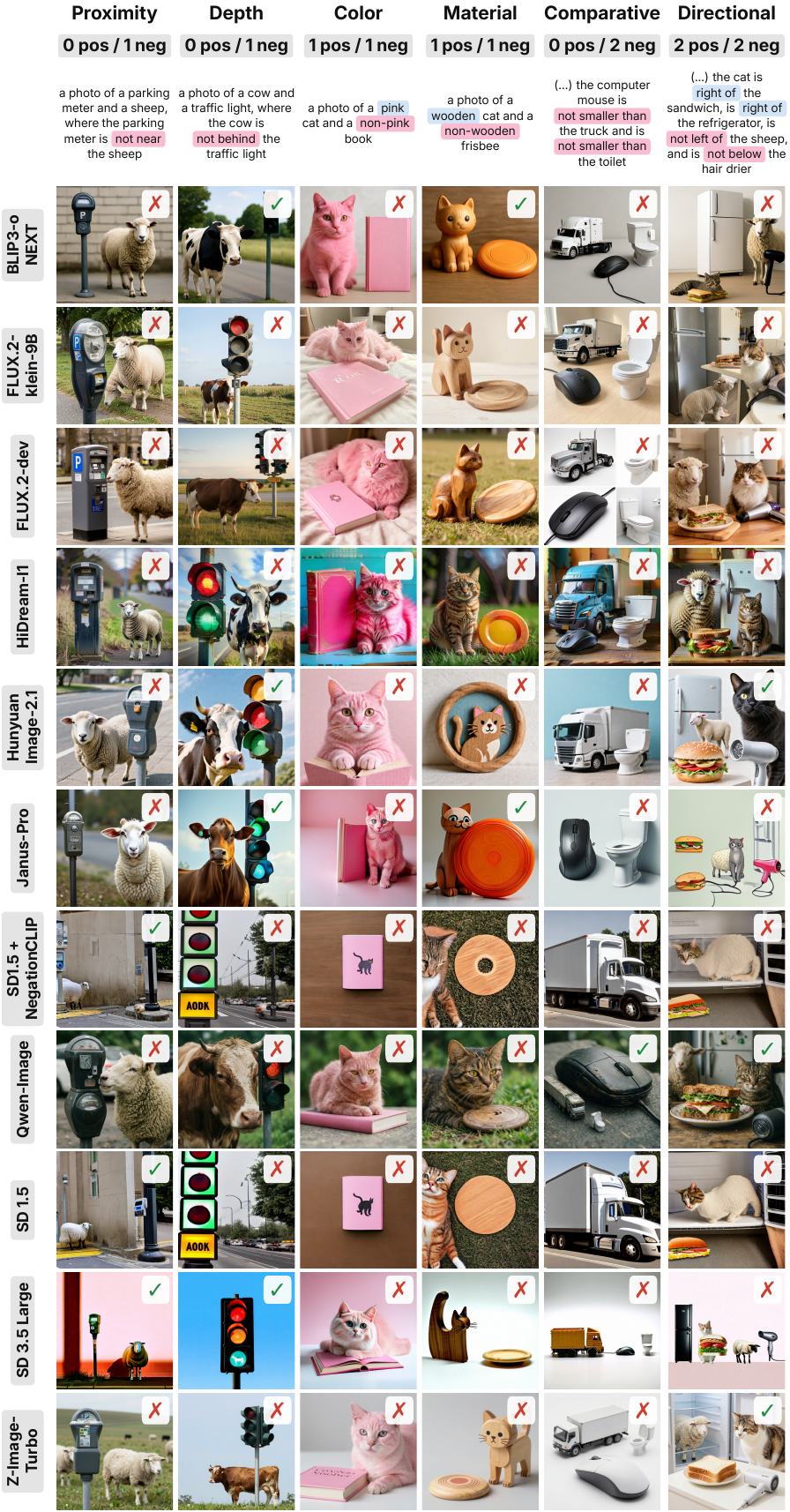}
 \caption{\textbf{Qualitative examples.} Six selected prompts (one per category) generated by all eleven T2I models.}

 \label{fig:qualitative-images-examples}
\end{figure*}

Figure~\ref{fig:qualitative-images-examples} presents selected generations
for six \bench prompts, one from each semantic category, across all eleven
evaluated T2I models. Showing each model's response to the same
prompt illustrates differences in rendering requested properties and
excluding forbidden ones while preserving the named objects. Each image
is also scored by our evaluator, with pass/fail marks indicating whether
it satisfies the complete prompt. These selected examples complement
the quantitative results by making the benchmark's requirements and
the models' successes and failures concrete.

\section{Limitations}
\label{app:limitations}

\bench is limited to what can be verified automatically in both polarities: color, material, and four spatial and size relations over 79 COCO classes. Contact, interactions such as holding or riding, and the negated presence of whole objects are not covered, and are natural extensions as detectors improve. Comparative relations also measure apparent image size rather than real-world size.
Negation is expressed in one English form (``non-\textit{value}'', ``is not \textit{relation}''), and relation prompts link one subject to every target. This keeps polarity comparisons controlled, but robustness to paraphrases, other languages, and richer relation structures is not tested; these can be added on the same evaluation pipeline.

\bench measures whether negated content is absent from the image, not whether the model understood the negation. Because negated statements are easier to satisfy by chance than positive ones, a model that ignores a negation can still pass some negated statements. A control prompt that omits the tested attribute or relation would separate the two, and fits the current design.
Scores depend on off-the-shelf perception models (RF-DETR, SAM~2, SigLIP~2, and Depth-Anything~V2), so benchmark accuracy is bounded by their quality; the modular design lets each component be replaced as stronger models appear. Finally, we evaluate open-weight models under their native inference settings rather than a matched compute budget; closed systems can be scored with the same protocol.

\end{document}